\documentclass{article}
\usepackage{arxiv}

\usepackage{geometry}
\usepackage{graphicx}
\usepackage[hidelinks]{hyperref}

\usepackage{amssymb}
\usepackage{bm}
\usepackage{lineno}
\usepackage{amsmath}

\usepackage{multirow}
\usepackage{float}
\usepackage{tabularx}
\usepackage{subcaption}
\usepackage{booktabs}
\usepackage{array,xcolor}

\usepackage{algorithm, algpseudocode}

\usepackage{cite}
\title{Neuroscience inspired scientific machine learning (Part-2): Variable spiking wavelet neural operator}
\author{
  Shailesh Garg  \\
  Department of Applied Mechanics\\
  Indian Institute of Technology Delhi\\
  Hauz Khas, New Delhi 110016, India. \\
  \texttt{shaileshgarg96@gmail.com} \\
  \And
  Souvik Chakraborty  \\
  Department of Applied Mechanics\\
  Yardi School of Artificial Intelligence (YScAI)\\
  Indian Institute of Technology Delhi\\
  Hauz Khas, New Delhi 110016, India. \\
  \texttt{souvik@am.iitd.ac.in}}
\begin{document}
\maketitle
\begin{abstract}
We propose, in this paper, a Variable Spiking Wavelet Neural Operator (VS-WNO), which aims to bridge the gap between theoretical and practical implementation of  Artificial Intelligence (AI) algorithms for mechanics applications. With recent developments like the introduction of neural operators, AI's potential for being used in mechanics applications has increased significantly. However, AI's immense energy and resource requirements are a hurdle in its practical field use case. The proposed VS-WNO is based on the principles of spiking neural networks, which have shown promise in reducing the energy requirements of the neural networks. This makes possible the use of such algorithms in edge computing. The proposed VS-WNO utilizes variable spiking neurons, which promote sparse communication, thus conserving energy, and its use is further supported by its ability to tackle regression tasks, often faced in the field of mechanics. Various examples dealing with partial differential equations, like Burger's equation, Allen Cahn's equation, and Darcy's equation, have been shown. Comparisons have been shown against wavelet neural operator utilizing leaky integrate and fire neurons (direct and encoded inputs) and vanilla wavelet neural operator utilizing artificial neurons. The results produced illustrate the ability of the proposed VS-WNO to converge to ground truth while promoting sparse communication.
\end{abstract}
\section{Introduction}\label{section: introduction}
Artificial intelligence (AI) \cite{russell2010artificial,ongsulee2017artificial}, specifically neural network \cite{choi2020introduction,dong2021survey} algorithms, has revolutionized the computing world. Its applications can now be found in varying fields like academia, entertainment, manufacturing, and others.
However, with all its benefits, neural network algorithms suffer from excessive resource and energy requirements. Its fundamental building block involves continuous activations, which fire indiscriminately, regardless of the potential contribution of its local output to the network's global output. This continuous neuron activity leads to increased computation load and, thus, increased energy consumption. For mechanics applications like active control of drones and robots, where energy is in limited supply, we require algorithms that can tackle complex tasks in an energy-efficient way. This is where Spiking Neural Networks (SNNs) \cite{yamazaki2022spiking,ghosh2009spiking} come into the picture. These are dubbed to be the next generation of AI algorithms, and its energy efficiency is its biggest appeal. They utilize spiking neurons \cite{long2010review,pfeiffer2018deep} instead of artificial neurons, which promote event-driven communication, thus introducing sparsity and conserving energy.

The inspiration for spiking neurons is drawn from biological neurons, which process huge amounts of information with a scanty energy budget by utilizing the same event-driven mechanism. SNNs, although they have existed for several decades now, exploration has in recent years gained pace, owing to the pressing need for sustainable technology and the development of supporting hardware like neuromorphic chips \cite{young2019review,davies2021advancing,zhu2020comprehensive}. They have mainly found its use in solving classification tasks \cite{rathi2020enabling,turkson2021classification,antelis2020spiking,dora2016development} and, in those, have shown accuracy at par with the existing Artificial Neural Networks (ANNs). Among the various spiking neuron models available in the literature, the most prevalent in literature is the Leaky-Integrate and Fire (LIF) neuron \cite{long2010review,burkitt2006review}. LIF neuron strikes a good balance between biological plausibility and ease of implementation and hence is widely used in SNN algorithms. Other more biologically plausible spiking neuron models like the Hodgkin-Huxley model \cite{hodgkin1952quantitative} are often not used as their energy and memory requirements are often more compared to LIF, owing to its complexity.

For engineering applications, we deal with regression tasks, and this is where, SNNs value in its current form diminishes significantly. Despite rich literature in classification tasks, literature concerning SNNs in regression tasks \cite{henkes2022spiking,gehrig2020event,zhang2022sms,kahana2022function,zhang2023artificial,gruning2014spiking} is scarce. Most of the works have been proposed in the past few years, and mostly, the results produced do not match the accuracy achieved when using ANNs. A recently proposed paper \cite{zhang2023artificial} shows promising results for regression tasks; however, it utilizes ANN to SNN converted networks \cite{liu2022spikeconverter,bu2023optimal} and not natively trained SNNs. ANN converted networks, while a viable option, are (i) approximations at best because of assumptions required during conversion and (ii) often require longer spike trains, resulting in reduced energy savings, thus defeating the primary purpose. Furthermore, since ANN is used in the training phase without modification, energy-saving benefits are not observed in this phase.

In this paper, our goal will be to transition regression algorithm, useful for mechanics applications, from an ANN framework to a Spiking framework. Specifically, we will be focusing on the recently proposed Wavelet Neural Operator (WNO) \cite{tripura2023wavelet}, which is an operator learning algorithm. Operator learning algorithms learn the mapping between two infinite-dimensional functional spaces. DeepONet \cite{lu2021learning}, an operator learning algorithm, was among the first operator learning algorithms to show excellent generalization capabilities in learning complex Partial Differential Equations (PDEs). This was further improved by the introduction of neural operators like Fourier neural operator \cite{li2020fourier} and WNO. WNO's architecture is inspired by Green's function, and it utilizes convolution layers and wavelet transformation within its architecture to achieve its goal. It has shown excellent performance in learning various PDEs and tackling tasks \cite{tripura2023wavelet,N2024116546,thakur2022multi,tripura2023waveletElasto,navaneeth2023waveformer} involving complex physical processes.

Now to bring the vanilla WNO to a spiking framework, we will first require a spiking neuron, which can tackle seamlessly, the regression tasks. To this end, we will use the Variable Spiking Neuron (VSN) introduced in Part-1 of this paper, which, while encouraging sparse communication, can deal with continuous activations. VSN is an amalgamation of the LIF neuron and the artificial neuron. Its propensity for sparse communication is derived from the LIF neuron's dynamics, while its ability to process continuous activations is derived from the artificial neuron's dynamics. Because of sparse communication, although there is a marginal loss in accuracy, the energy savings achieved make it suitable for the next-generation AI algorithms.
The proposed spiking WNO will be utilizing the VSNs within its architecture to reduce the computational load in the forward pass of the neural network, thus conserving energy. The same will be referred to as the Variable Spiking WNO (VS-WNO) in the following text. The VSNs will primarily be used as activations in the proposed VS-WNO, but the same may also be used to introduce sparsity in the input data. We also propose a spiking loss function, which promotes even more sparse communication by placing restraints on the spiking activity of the network. For training the proposed VS-WNO, surrogate backpropagation \cite{neftci2019surrogate,eshraghian2023training} is used. 

The proposed VS-WNO is tested against both one-dimensional and two-dimensional PDEs, and the results produced show good convergence to the ground truth while achieving sparse communication. Comparisons of the results have been drawn against those produced using vanilla WNO and those produced using WNO, utilizing LIF neurons. The rest of the paper is arranged as follows, Section \ref{section: background} discusses the background of WNO and spiking neurons. Section \ref{section: proposed framework} discusses the proposed VS-WNO and the proposed spiking loss function. Section \ref{section: numerical} discusses the various examples, and Section \ref{section: conclusion} concludes the findings of the manuscript. 
\section{Background}\label{section: background}
\subsection{Wavelet neural operator:} 
\begin{figure}[ht!]
    \centering \includegraphics[width = 0.8\textwidth]{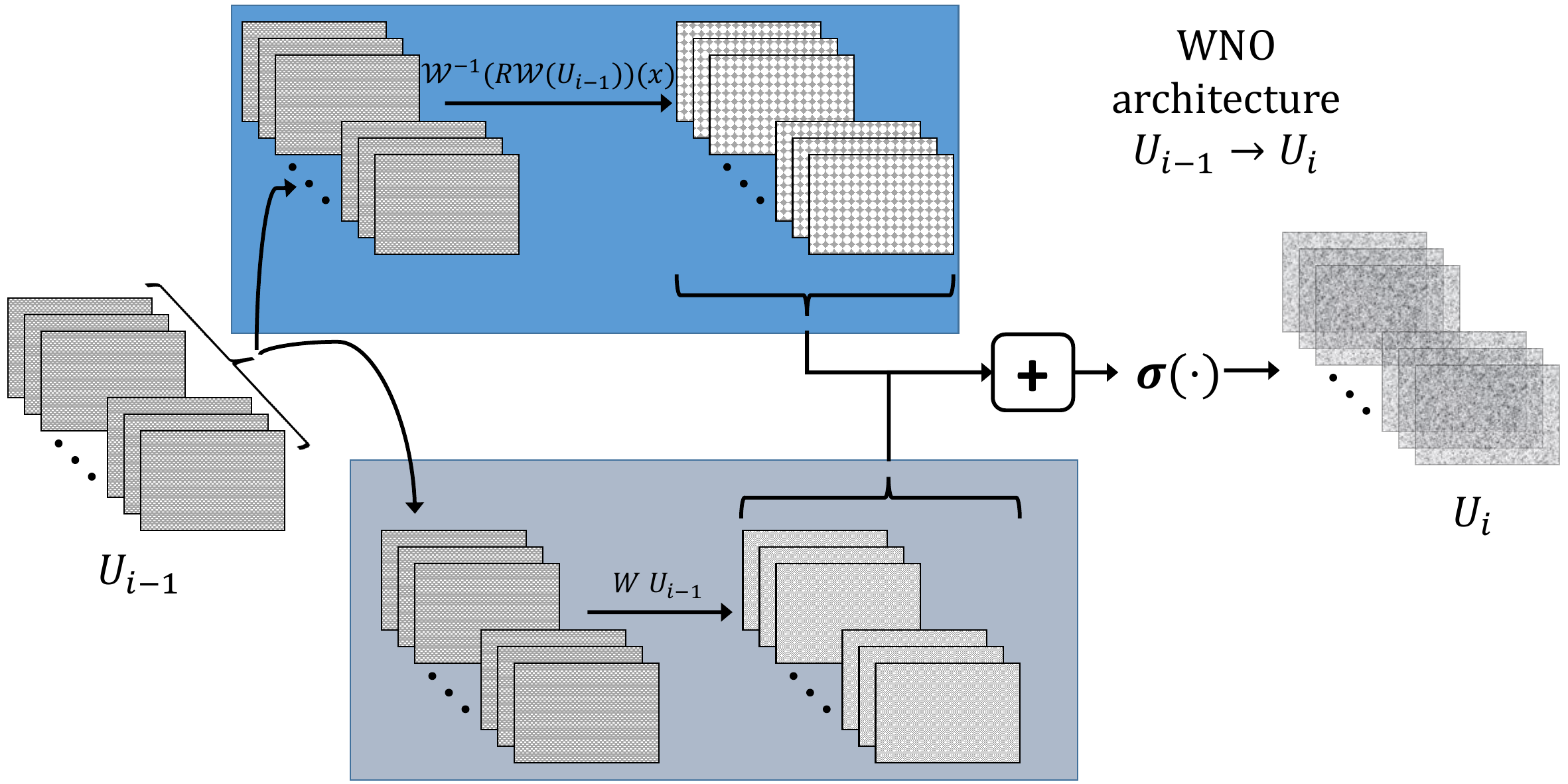}
    \caption{Information flow in vanilla wavelet neural operator.}
    \label{fig: wno ui-1 to ui}
\end{figure}

Neural operators are a class of neural network algorithms that aim to learn the mapping between two infinite-dimensional functional spaces. For a domain $D\in\mathbb R^d$ discretized at $n$-points and $x\in D$, consider a PDE for which we have observations of input function $I\in\mathbb R^{n\times d_a}$ and output function $\widetilde O\in\mathbb R^{n\times d_o}$. Neural operators are inspired by Green's function, and through the use of convolution operations, they learn the operator $\mathcal D: I\times\bm\theta\rightarrow\widetilde O$. $\bm\theta$ here represents the learnable parameters of the neural network. The input $I(x)$ is first uplifted to $U_0\in\mathbb R^{d_u}$ and then $L$ iterative updates $U_i = \mathcal F(U_{i-1})$ are applied, where $j = 1,2,3,...,L$. The updated state, $U_i\in\mathbb R^{d_u}$ is computed as follows,
\begin{equation}
    U_i = \kappa\left(\mathcal W^{-1}(R\,\mathcal W(U_{i-1}))(x)+W U_{i-1}\right),
    \label{eqn: ui}
\end{equation}
where $\kappa(\cdot)$ is an activation function, $W:\mathbb R^{d_u}\rightarrow \mathbb R^{d_u}$ is a linear transformation and $R$ is a parameterized kernel. $\mathcal W(\cdot)$ and $\mathcal W^{-1}(\cdot)$ in Eq. \eqref{eqn: ui}, represents the wavelet and inverse wavelet transform and for a function $f:D\rightarrow\in\mathbb R^{d_u}$, the same are defined as follows,
\begin{equation}
\begin{gathered}
\mathcal W(f)(x) = f_w(s,t_r) = \int_D f(x)\dfrac{1}{|s|^{1/2}}\psi\left(\dfrac{x-t_r}{s}\right)dx\\
\mathcal W^{-1}(f_w)(s,t_r) = f(x) = \dfrac{1}{C_\psi}\int_0^\infty\int_D f_w(s,t_r)\dfrac{1}{|s|^{1/2}}\tilde\psi\left(\dfrac{x-t_r}{s}\right)dt_r\dfrac{ds}{s^2},
\end{gathered}
\label{eq: w iw transform}
\end{equation}
where $\psi(x)\in\mathcal L^2(\mathbb R)$ is an orthonormal mother wavelet that is localized in both the time and frequency domain. $\tilde \psi$ is the dual function of mother wavelet $\psi$. $s\in \mathbb R^+$ and $t_r\in \mathbb R$ are the scaling and translational parameters. $C_\psi\in\mathbb R^+$ is called the admissible constant and is given as follows,
\begin{equation}
    C_\psi = 2\pi\int_D\dfrac{|\psi_\omega|^2}{|\omega|}d\omega,
\end{equation}
where $\psi_\omega$ is defined as the fourier transform of $\psi$. A schematic for the information flow from $U_{i-1}$ to $U_i$ is given in Fig. \ref{fig: wno ui-1 to ui}. The final output $\widetilde O$ is obtained by applying transformation $\mathcal G: \mathbb R^{d_u}\rightarrow\mathbb R^{d_o}$ on $U_L$ i.e. $\widetilde O=\mathcal G(U_L)$. For more details on the vanilla WNO, readers may follow \cite{tripura2023wavelet}.

\subsection{Neuron models} 
Deep learning algorithms emulate the human nervous system and consist of a web of interconnected neurons. These neurons are meant to represent the biological neurons, which receive inputs from various sources, accumulate them, and pass them forward only when a certain threshold is crossed. In deep learning, however, the most popular neuron model is the artificial neuron, wherein the inputs are passed through an activation function, and we receive output for every input regardless of its contribution to the final output. For a given input $z$, the output $y$ from an artificial neuron can be computed using activation function $\sigma(\cdot)$ as $y = \sigma(z)$. $z$ itself can be obtained using a suitable transformation, for example, a convolution operation. 

The artificial neuron equipped artificial neural networks have shown excellent performance in tackling a multitude of tasks. However, the artificial neuron's indiscriminate firing increases the overall network computational load, thereby increasing ANN's energy requirements. To remedy this, researchers are once again looking towards biological neurons. The motivation for this is the low-energy information processing in the brain and human nervous system \cite{furber2016brain}. This is attributed to the sparsity in communication observed in the human nervous system. The sparsity is a direct result of the all-or-none mechanism of biological neurons, wherein the information coming from various sources across several spike time steps\footnote{Spike time steps: Information in the biological neurons and the spiking neurons is transferred in the form of spikes. $n_s$ spikes representing one encoded packet of information are referred to as a spike train, and Spike Time Steps (STSs) refer to individual events in this spike train. Event here, implying weather spike or no spike.} is stored and is transferred to subsequent neurons only when a set threshold is crossed. While several mathematical models for spiking neurons exist in the literature, the LIF model is among the most popular. The dynamics of the LIF model are described as follows,
\begin{equation}
\begin{aligned}
    M^{(\bar t)} &= \beta M^{(\bar t-1)} + z^{(\bar t)},\\
    y^{(\bar t)} &= \left\{\begin{array}{ll}
    1;& \,\,\,\,\, M^{(\bar t)}\ge T\\
    0; & \,\,\,\,\, M^{(\bar t)}<T\end{array}\right.,
    \text{\,\,\,\,\,if } y^{(\bar t)} = 1, M^{(\bar t)} \leftarrow 0,
\end{aligned}
\label{equation: LIF eqns}
\end{equation}
where $z^{(\bar t)}$ represents the input at $\bar t$ STS and $M^{(\bar t)}$ represents the memory of LIF neuron at $\bar t$ STS. $\beta$ is the leakage parameter, controlling the influence of past memory, and $T$ is the threshold against which the information stored is tested. $y^{(\bar t)}$ is the output at $\bar t$ STS. The final output $y = \{y^{(1)}, y^{(2)}, y^{(3)},\ldots, y^{(\bar T)}\}$, where $\bar T$ is the number of STS in the spike train. 
Henceforth, the terminology \textit{spiking neural networks} will be used to refer to networks that utilize LIF neurons.

In addition to the spiking neurons, another important facet of SNNs is the input encoding. As discussed previously, the information transfer in SNNs is in the form of spikes. The multiple STS creates an opportunity for sending the input data after encoding \cite{auge2021survey,kim2022rate} it into spikes. The encoding can be achieved by various means,
however, in the prevailing literature, rate encoding is among the most popular choice. The goal is to match the average number of spikes in the spike train with the quantity being encoded. Rate encoding, although suitable for classification tasks, for regression, since mapping is between real numbers, any information loss can be detrimental to final results. Now rate encoding can potentially retain information to a high degree of precision, but the required number of spike time steps will be large, leading to increased spiking activity overall, thus defeating our original goal of energy consumption. 

For regression tasks, \cite{zhang2022sms} introduces a triangular encoding. In this, each spike is given a specific value equal to the inverse of the total STS, and the input quantity is encoded accordingly. For example, a quantity 0.7 to be encoded using 10 spikes will have seven spikes followed by three no-spike events, represented as $[1 1 1 1 1 1 1 0 0 0]$. In this encoding, spikes are continuously observed till the quantity under consideration is approximated, and then no-spike events are observed till the last STS. The triangular encoding is defined for one-dimensional functions. Another alternative is to give direct inputs \cite{kim2022rate}, repeated at all spike time steps. This allows for retaining the information with the maximum precision possible.
\section{Variable spiking WNO}\label{section: proposed framework}
The proposed VS-WNO has three main components, which need to be defined, (i) the variable spiking neuron, (ii) the VS-WNO architecture, and (iii) the encoding technique used. We will discuss, in this section, each component in the sequence of its introduction above. 
\subsection{Variable spiking neurons:} 
\begin{figure}[ht!]
    \centering
    \includegraphics[width = 0.95\textwidth]{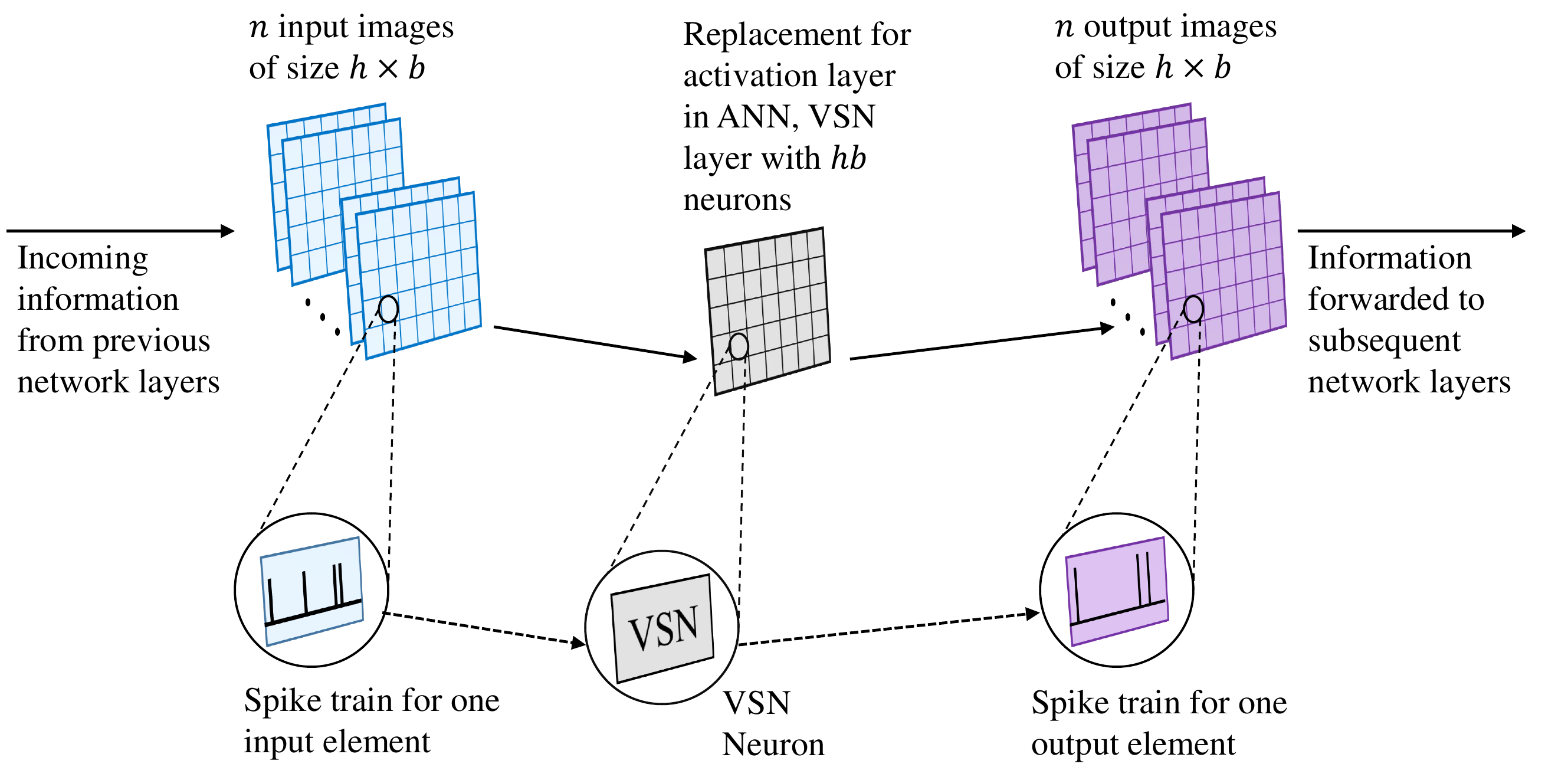}
    \caption{Placement of VSN layer in an artificial neural network.}
    \label{fig: VSN placement}
\end{figure}

Spiking neurons, discussed previously, have shown great promise in tackling classification tasks, but their performance in regression tasks is middling at best. To remedy this, 
the authors introduced Variable spiking neurons in the Part-1 of this paper, 
which are an amalgamation of LIF spiking neurons and artificial neurons. The idea behind VSN is to send a graded spike in place of the binary spike in the event that the memory of the neuron crosses the set threshold. The graded spike can be equal to the input of the neuron or input passed through a continuous activation. The dynamics of VSN is defined as follows,
\begin{equation}
\begin{aligned}
    M^{(\bar t)} &= \beta M^{(\bar t-1)} + z^{(\bar t)},\\
    \widetilde y^{(\bar t)} &= \left\{\begin{array}{ll}
    1;&\,\,\,\,\, M^{(\bar t)}\ge T\\
    0;&\,\,\,\,\, M^{(\bar t)}<T\end{array}\right.,\text{\,\,\,\,\,if } \tilde y^{(\bar t)} = 1, M^{(\bar t)}  \leftarrow 0\\
    y^{(\bar t)} &= \sigma(z^{(\bar t)}\tilde y^{(\bar t)}), \,\,\,\,\,\text{given, }\sigma(0) = 0, 
\end{aligned}
\label{equation: VSN eqns}
\end{equation}
In the event, no spike is observed at $\widetilde y^{(\bar t)}$, $(z^{(\bar t)}\tilde y^{(\bar t)}) = 0$ and thus no information should flow forward. The constraint, $\sigma(0) = 0$, ensures this, thus promoting sparsity. $\sigma(\cdot)$ can be a linear or nonlinear continuous function. Henceforth, the networks utilizing VSNs will be termed as Variable SNNs (VSNNs). The laver containing VSNs replaces the activations in an artificial neural network. A schematic for the placement of VSN is shown in Fig. \ref{fig: VSN placement}.
\subsubsection{Energy consumption of VSN}
In any neural network, energy $E$ is spent on three main operations, (i) retrieving parameters, (ii) synaptic operations, i.e. computations of $z$, and (iii) broadcasting activation.
\begin{equation}
    E = E_{rp}+E_{so}+E_{a},
\end{equation}
where $E_{rp}$ is the energy spent in retrieving parameters, $E_{so}$ is the energy spent in synaptic operations and $E_{a}$ is the energy spent in broadcasting activation. Now, the energy spent in these operations is vastly dependent on the implementation and optimizations done to both hardware and software. The manuscript, \cite{davidson2021comparison}, shows that within synaptic operations, depending on the type of operations, multiple subdivisions can be made. For convolution operation, the four sub-operations involved are retrieving neuron state, multiplication with weight, addition to neuron state, and storing neuron state. Based on energy estimation data from post-layout analysis of SpiNNaker2, the energy consumed in retrieving neuron state and in multiplication with weight is taken to be 5$\mathcal E$. $\mathcal E$ represents the energy consumed in the addition operation. The energy consumed in the storing neuron state is also equal to $\mathcal E$. Thus an artificial neuron will consume an energy of 12$\mathcal E\times N_{\text{mt}}$ and a spiking neuron with binary spikes will consume 7$\mathcal E\times N_{\text{mt}}\times N_{s}$. $N_s$ here represents the average number of spikes, and $N_{\text{mt}}$ here represents the target nodes for the current node under consideration. The reduction in energy consumed, observed in spiking neurons, is because of the elimination of multiplication operation, and spiking neurons will be energy efficient if $N_s$ across all STS is less than 1.7. Now, extending this analysis to variable spiking neurons, the energy consumed will be equal to 12$\mathcal E\times N_{\text{mt}}\times N_{s}$ because of non-binary spikes and reintroduction of the multiplication operation. However, the VSN will conserve energy if the $N_s$ is less than 1.0. 



In the literature \cite{davidson2021comparison,dampfhoffer2022snns,lemaire2022analytical,zhang2023artificial}, it is well established that the energy spent is dependent on the spiking activity or firing rate of a neuron. More is the spiking activity, the more computations will be required, and hence more will be the energy consumed. The vice-versa is true as well.
Following the trend from the literature, in the subsequent numerical illustrations, we will discuss the percentage average spiking activity of the VSN neuron in lieu of energy metric. To compute the percentage average spiking activity, we use a metric, \textit{average percentage spikes}, $\mathcal S$, which is computed as follows,
\begin{equation}
    \mathcal S = 100\,\,\dfrac{\text{number of spike produced in VSN layer across all STS}}{\text{total possible spikes per STS} \times \text{number of STS}}.
\end{equation}
\subsubsection{Spiking loss function}
The leakage and threshold parameters of the VSN need to be initialized and can be treated as either fixed hyperparameters or tuned by treating them as trainable parameters. By treating them as fixed, we can run into convergence issues, leading to a loss in accuracy. By training them, we can achieve great accuracy, but a drawback here is that it is possible during training these parameters, along with other trainable parameters like weights and bias, are getting values that prioritize accuracy over sparsity. This can lead to reduced energy savings, thus defeating the initial purpose of using spiking neurons. To avoid this situation, we propose a new loss function, which places the restriction on the spiking activity, thus ensuring sparse communication. 

As discussed previously, in a variable spiking neural network, the VSN layer is placed as an activation layer. Say there are $n$ VSN layers $A_{1:n}$, each having average percentage spikes $\mathcal S_{1:n}$.
We define a total network average spike $\widetilde{\mathcal S} = \sum_i \mathcal S_i/100$. Now, to constraint this spiking activity, despite learnable VSN parameters, we introduce a spiking loss function $L_s$, computed as follows,
\begin{equation}
    L_s = \alpha L_b+ \gamma \widetilde{\mathcal S},
\end{equation}
where $L_b$ represents the basic loss function to train any neural network like mean square error or $L^2$ error. $\alpha$ and $\gamma$ are weights assigned to $L_b$ and $\widetilde{\mathcal S}$ respectively and are tunable hyperparameters. It should be noted here that because of reduced spiking activity, the accuracy achieved may be reduced marginally, but by tuning the parameters $\alpha$ and $\gamma$, we can strike a balance between energy efficiency and reduced accuracy. Through empirical evidence, shown in numerical illustrations, it is observed that the spiking loss function is able to reduce the spiking activity even for cases where the VSN parameters are treated as fixed. The hypothesis here is that the remaining trainable parameters of the network architecture are tuned to minimize spiking activity, thus optimizing the loss function.
\subsection{VS-WNO architecture:} 
\begin{figure}[ht!]
    \centering \includegraphics[width = 0.8\textwidth]{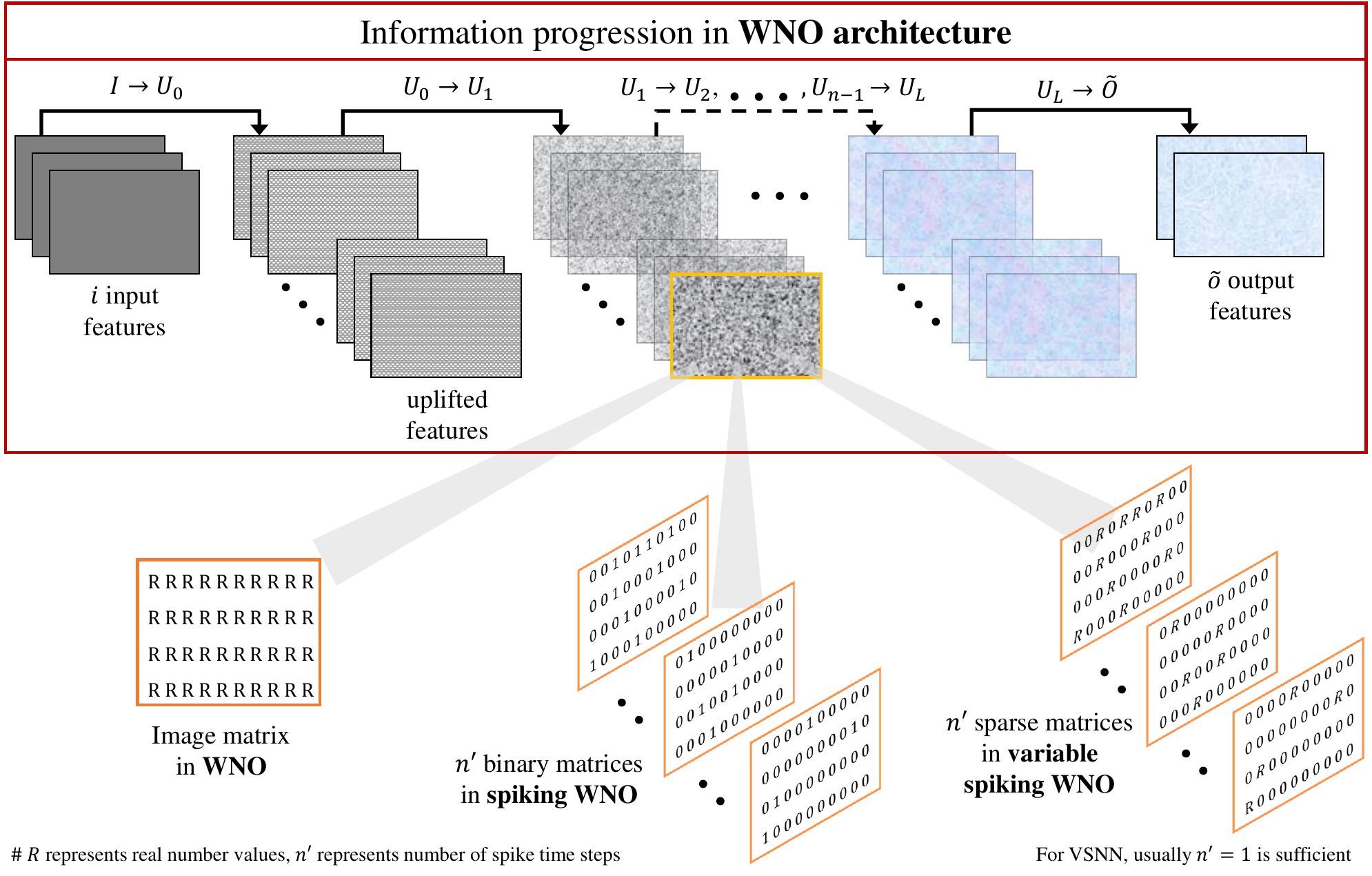}
    \caption{Quality of information transferred in vanilla WNO, spiking WNO (utilizing LIF neurons), and VS-WNO}
    \label{fig: information various WNOs}
\end{figure}

The architecture for vanilla WNO has one uplifting transition, $\mathcal H:\mathbb R^{d_a}\rightarrow\mathbb R^{d_u}$, which uplifts the input $I$ to $U_0 = \mathcal H(I)$, $L$ recursive update layers, $U_{1:L}$, and another transition $\mathcal G:\mathbb R^{d_u}\rightarrow\mathbb R^{d_o}$ to bring the dimensions of the output of $U_L$ layer to the desired output size.
The activations in vanilla WNO are placed on the update layers and on the layers of the Forward Neural Network (FNN) representing the transition $\mathcal G$. Usually, GeLU activation is used within the vanilla WNO architecture. However, it should be noted here that there are no theoretical restrictions on the placement or type of activations used.
Now, the basic architecture of the proposed VS-WNO is the same as the vanilla WNO, the key difference being that the VS-WNO utilizes VSNs within its architecture in order to promote sparse communication and consequently conserve energy. The spiking neurons replace the activations of vanilla WNO. The $\sigma(\cdot)$ within the VSN architecture can be GeLU activation, the popular choice of activation in a vanilla WNO, or it may be a linear activation, depending on the dataset being learned. The linear activation will result in greater energy savings because of reduced computations, and the same produce competitive results, as shown in numerical illustrations. A schematic showing the difference in the quality of information transferred in vanilla WNO, spiking WNO (utilizing LIF neurons), and VS-WNO is shown in Fig. \ref{fig: information various WNOs}. In the spiking WNO or VS-WNO, information is sparse, as shown in the visual representation.
Algorithm \ref{algo: VS WNO} details the steps for training the proposed VS-WNO.
\begin{algorithm}[ht!]
\caption{Training algorithm for the VS-WNO architecture.}
\label{algo: VS WNO}
\begin{algorithmic}[1]
\renewcommand{\algorithmicrequire}{\textbf{Initialize:}}
\Require{Initialize network parameters.}
\For{i = 1\,:\,epochs}
\State Initialize memory for various VSNs within the architecture.
\State Initialize an empty matrix $\mathbb E$ to store output obtained at various STS.
\For{j = 1\,:\,no. of STS}
\State Pass inputs through the FNN representing the transition $\mathcal H$ and obtain $U_0$.
\State Pass $U_0$ through $L$ update layers $U_{0:L}$.
\State Pass output of $U_L$ through all but the last layer of the FNN representing transition $\mathcal G$.
\State Store the output in the matrix $\mathbb E$.
\EndFor
\State Take the mean of matrix $\mathbb E$ along the STS dimension.
\State Pass the mean through the last layer of FNN representing transition $\mathcal G$.
\State Compare with ground truth and compute objective function (loss function).
\State Update network parameters.
\EndFor
\renewcommand{\algorithmicrequire}{\textbf{Output:}}
\Require{Trained VS-WNO model.}
\end{algorithmic}
\end{algorithm}
\subsubsection{Encoding}
The data being sent into the VS-WNO is not encoded but rather sent directly. If the STS required for convergence to ground truth is more than one, each STS will receive the same input. As discussed, direct encoding is preferred for regression tasks; since the mapping is between real numbers, we want to retain the maximum precision possible. Any encoding technique, depending on the number of STS used, will lose some information, which can lead to erroneous results. For example, if the number 0.1234 is to be encoded using triangular encoding having 10 STS, information only up to the first decimal place will be retained. Because sparsity in communication is the main goal of the spiking neural networks, having a large number of STS is counterintuitive and can result in increased energy consumption, thus defeating the main goal.
\section{Numerical Illustrations}\label{section: numerical}
In this section, we discuss four examples to test the proposed VS-WNO, covering both one-dimensional and two-dimensional PDEs. The first example deals with the one-dimensional Burgers' equation, and the second example covers the Allen-Cahn equation. The third and fourth examples cover Darcy's equation solved on rectangular and triangular domains, respectively. To train the Vanilla WNO, the backpropagation algorithm is used, whereas to train the spiking WNO and VS-WNO, surrogate backpropagation is used. In surrogate backpropagation, during backward pass, the threshold function of the spiking neuron or VSN is idealized using a fast sigmoid \cite{zenke2018superspike,eshraghian2023training} function with a slope parameter equal to 25. The nomenclature for various neural networks implemented in the following section is given in Table \ref{tab: nomenclature}.
\begin{table}[ht!]
    \caption{Description of various networks used in the examples.}
    \vspace{1em}
    \centering
    \begin{tabular}{p{0.20\textwidth}p{0.75\textwidth}}
    \toprule
    Name & Description\\
    \midrule
    WNO & Vanilla WNO utilizing artificial neuron and GeLU activation, wherever required \\
    LIF-WNO & Spiking WNO utilizing LIF neurons and receiving direct inputs without encoding, 1 STS\\
    LIF-WNO-RE-10 & Spiking WNO utilizing LIF neurons and receiving rate encoded inputs with 10 STS\\
    LIF-WNO-TE-10 & Spiking WNO utilizing LIF neurons and receiving triangular encoded inputs with 10 STS\\
    VS-WNO-L & Variable spiking WNO utilizing VSN neurons with linear activation and receiving direct inputs, 1 STS\\
    VS-WNO-GeLU & Variable spiking WNO utilizing VSN neurons with GeLU activation and receiving direct inputs, 1 STS\\
    \bottomrule
    \end{tabular}
    \label{tab: nomenclature}
\end{table}

The details for architectures of various networks are as follows. The transitions $\mathcal H$ and $\mathcal G$ are carried out using fully connected neural networks with one and two layers, respectively. The details of nodes in fully connected layers of the transition layer, the number of recursive update layers, the mother wavelet used for wavelet transform, and the number of wavelet decompositions $m$ in each update layer are given in Table \ref{tab: wno details}.
\begin{table}[ht!]
    \caption{Network details for various examples. db\# here represents Daubechies wavelets with \# number of wavelet and scaling function coefficients.}
    \vspace{1em}
    \centering
    \begin{tabular}{lcccccccc}
    \toprule
    \multirow{2}{*}{Example} & \multirow{2}{*}{$L$} & \multirow{2}{*}{Wavelet} & \multirow{2}{*}{$m$} & \multicolumn{2}{c}{Nodes in first layer} && \multicolumn{2}{c}{Number of samples}\\\cmidrule{5-6}\cmidrule{8-9}
    & & & & $\mathcal H$ & $\mathcal G$ && Training & Testing\\
    \midrule
    Burgers & $4$ & db6 & $8$ & 64 & 128 && 1000 & 100\\
    Allen Cahn & $4$ & db4 & $1$ & 64 & 128 && 1400 & 100\\
    Darcy: rectangular grid & $4$ & db4 & $4$ & 64 & 128 && 1000 & 100\\
    Darcy: triangular grid & $4$ & db6 & $3$ & 64 & 128 && 1900 & 100\\
    \bottomrule
    \end{tabular}
    \label{tab: wno details}
\end{table}
Note that the second layer of the fully connected neural network, representing $\mathcal G$, has a single node in all examples. In vanilla WNO, GeLU activation is used after $U_1$, $U_2$, $U_3$, and the first layer of the network representing $\mathcal G$. In spiking WNO networks, the GeLU is replaced by a layer containing LIF neurons, and in VS-WNO, GeLU is replaced by a layer containing VSN neurons. The number of spiking neurons will depend on the input signal shape for all these layers. In the recursive update layers $U_i$, the quantity $WU_{i-1}$ is represented by a convolution layer having a kernel size of one. All networks are trained for 500 epochs. The learning rate is taken as $10^{-3}$ for various networks unless otherwise mentioned.
ADAM optimizer is used for training, with a weight decay of $10^{-4}$ for Burgers' and both Darcy examples, and is taken equal to $10^{-6}$ for Allen Cahn equation. For further details, please refer to the GitHub repository \footnote{Link for GitHub repository shall be provided upon acceptance}. Unless mentioned otherwise, the parameters of LIF neurons and VSNs are considered trainable.

The percentage normalized $L^2$ error, $\epsilon$, and average percentage spikes, $\mathcal S_i$, after each activation layer $A_i$, reported below for various networks are produced after running the same network five times. The reported values are the mean and standard deviation of collective results from these five trials. The plots shown below in various examples for various networks are produced using only one of these five trials. It should be noted that the results produced using vanilla WNO are considered the gold standard, and the goal of the proposed VS-WNO will be to achieve similar/surpass the performance of vanilla WNO.
\subsection{Example 1: Burgers' equation}
The one-dimensional Burgers' equation is defined as follows,
\begin{equation}
    \begin{gathered}
    \dfrac{\partial u(x,t)}{\partial t}+u(x,t)\dfrac{\partial u(x,t)}{\partial x} = \nu\dfrac{\partial^2 u(x,t)}{\partial x^2},\\
    x\in(0,1],\,\,\,\,\,t\in[0,1],
    \end{gathered}
    \label{eqn: Burgers}
\end{equation}
where $u(x,t)$ is the velocity and $\nu = 0.1$ is the viscosity term. Periodic boundary conditions, $u(0,t)=u(1,t)$ are considered and the initial condition, $u(x,0)$ is sampled from a Gaussian random field $\mathcal N(0,625(-\Delta+25\,\mathbb I)^{-2})$. The goal here is to learn the mapping between $u(x,0)$, i.e. the initial condition, and $u(x,1)$, i.e. solution at $t=1$. The datasets are taken from \cite{li2020fourier}, and the domain is divided into 1024 discretizations. A thousand samples were used for training the various networks. The network details are as discussed previously, with changes in the learning rate for LIF-WNO, LIF-WNO-RE-10, and LIF-WNO-TE-10 networks. For these the learning rate is taken as $10^{-4}$, $10^{-4}$, and $5\times10^{-5}$, respectively.

Table \ref{tab: initial results all examples} shows the percentage normalized $L^2$ error values observed for various networks when trained for Burgers' equation. Results for the Burgers' equation were also generated using LIF-WNO-TE-10, and a percentage error $\epsilon$ value of $40.23\pm 1.57$ was observed for test dataset predictions. As can be seen, the VS-WNO (both with GeLU and linear activation) outperforms all LIF-WNO networks, and the error observed is comparable to that observed in the vanilla WNO network.
\begin{table}[ht!]
    \caption{The percentage error $\epsilon$ observed in various examples. Predictions are made for test datasets using different networks. The values are reported in the format, $\underset{\pm\text{ std. dev.}}{\text{mean}}$. The mean and standard deviations are of results produced in five different runs of the networks.}
    \vspace{1em}
    \centering
    \begin{tabular}{lccccc}
    \toprule\cline{2-2}
    Example & \multicolumn{1}{|c|}{WNO} & LIF-WNO & LIF-WNO-RE-10 & VS-WNO-L & VS-WNO-GeLU\\\cline{2-2}\\[-1em]\midrule\cline{2-2}
    
    Burgers' & \multicolumn{1}{|c|}{$\underset{\pm 0.03}{1.88}$} & $\underset{\pm 1.51}{24.25}$ & $\underset{\pm 3.09}{45.83}$ & $\underset{\pm 0.10}{2.87}$ & $\underset{\pm 0.19}{2.70}$\\
    
    Allen Cahn & \multicolumn{1}{|c|}{$\underset{\pm 0.004}{0.23}$} & $\underset{\pm 0.19}{4.30}$ & $\underset{\pm 0.26}{65.44}$ & $\underset{\pm 0.10}{1.11}$ & $\underset{\pm 0.19}{1.10}$\\
    
    Darcy Recatangular & \multicolumn{1}{|c|}{$\underset{\pm 0.06}{1.77}$} & $\underset{\pm 0.41}{8.94}$ & $\underset{\pm 7.10}{12.61}$ & $\underset{\pm 0.02}{1.81}$ & $\underset{\pm 0.06}{1.82}$\\
    
    Darcy Triangular & \multicolumn{1}{|c|}{$\underset{\pm 0.02}{0.88}$} & $\underset{\pm 0.65}{15.52}$ & $\underset{\pm 3.42}{42.39}$ & $\underset{\pm 0.07}{0.97}$ & $\underset{\pm 0.03}{0.80}$\\[0.5em]
    \cline{2-2}\\[-1em]
    \bottomrule
    \end{tabular}
    \label{tab: initial results all examples}
\end{table}

\begin{figure}[ht!]
    \centering
    \includegraphics[width = 0.80\textwidth]{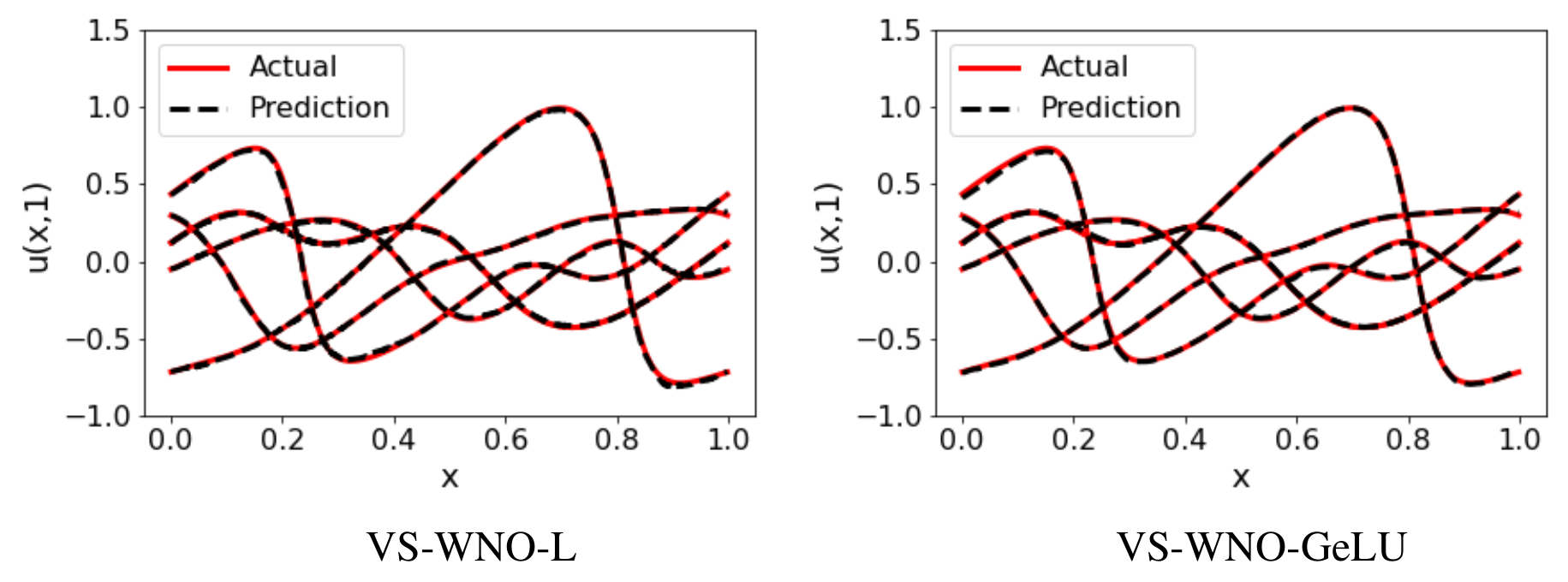}
    \caption{Predictions (five samples) using VS-WNO-L and VS-WNO-GeLU networks, compared against the ground truth for Burger's example.}
    \label{fig: Burgers' VS-WNO OL}
\end{figure}

\begin{figure}[ht!]
\centering
\begin{subfigure}{0.475\textwidth}
    \centering
    \includegraphics[width = \textwidth]{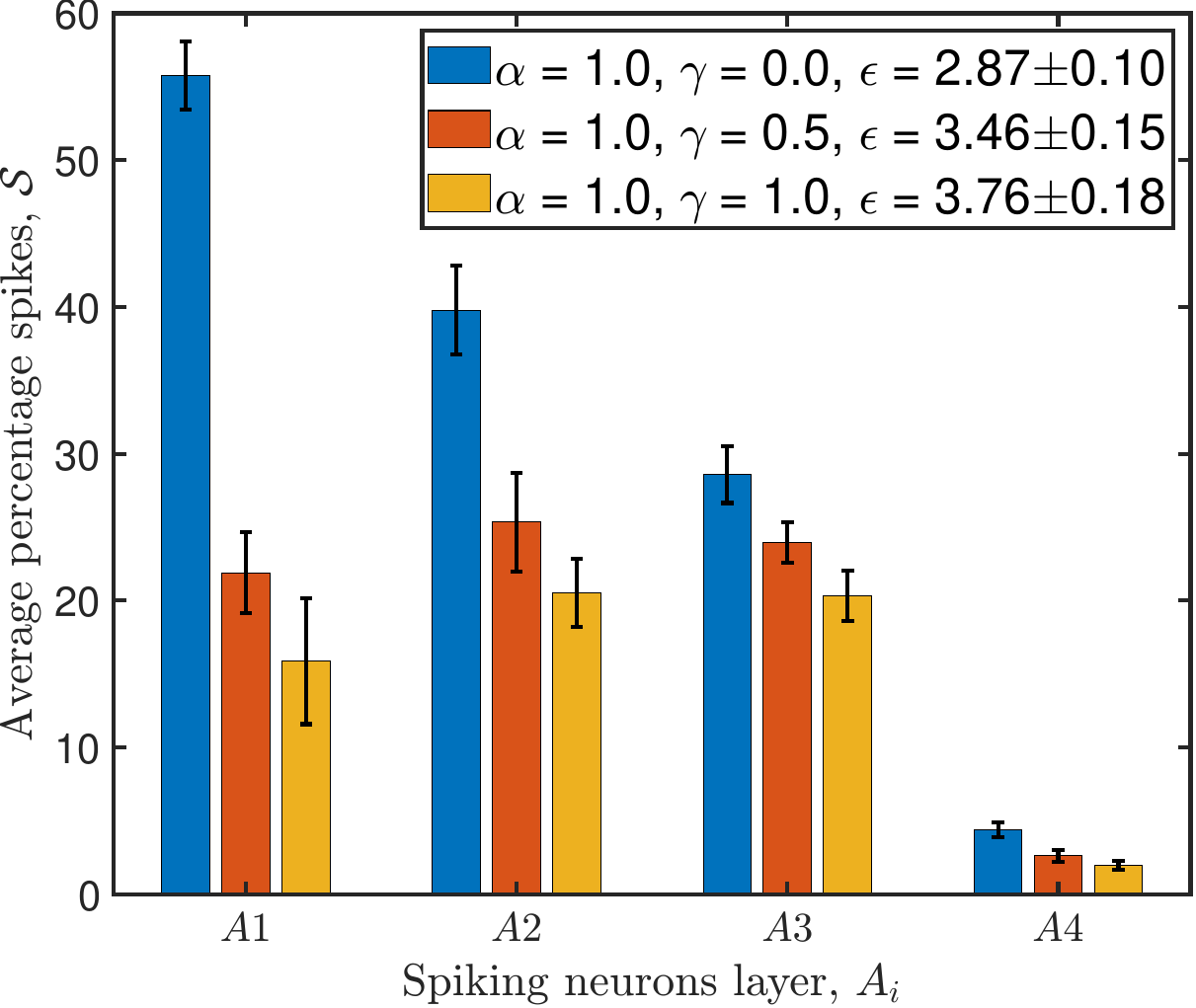}
    \caption{VS-WNO with Linear activation}
\end{subfigure}
\hspace{1em}
\begin{subfigure}{0.475\textwidth}
    \centering
    \includegraphics[width = \textwidth]{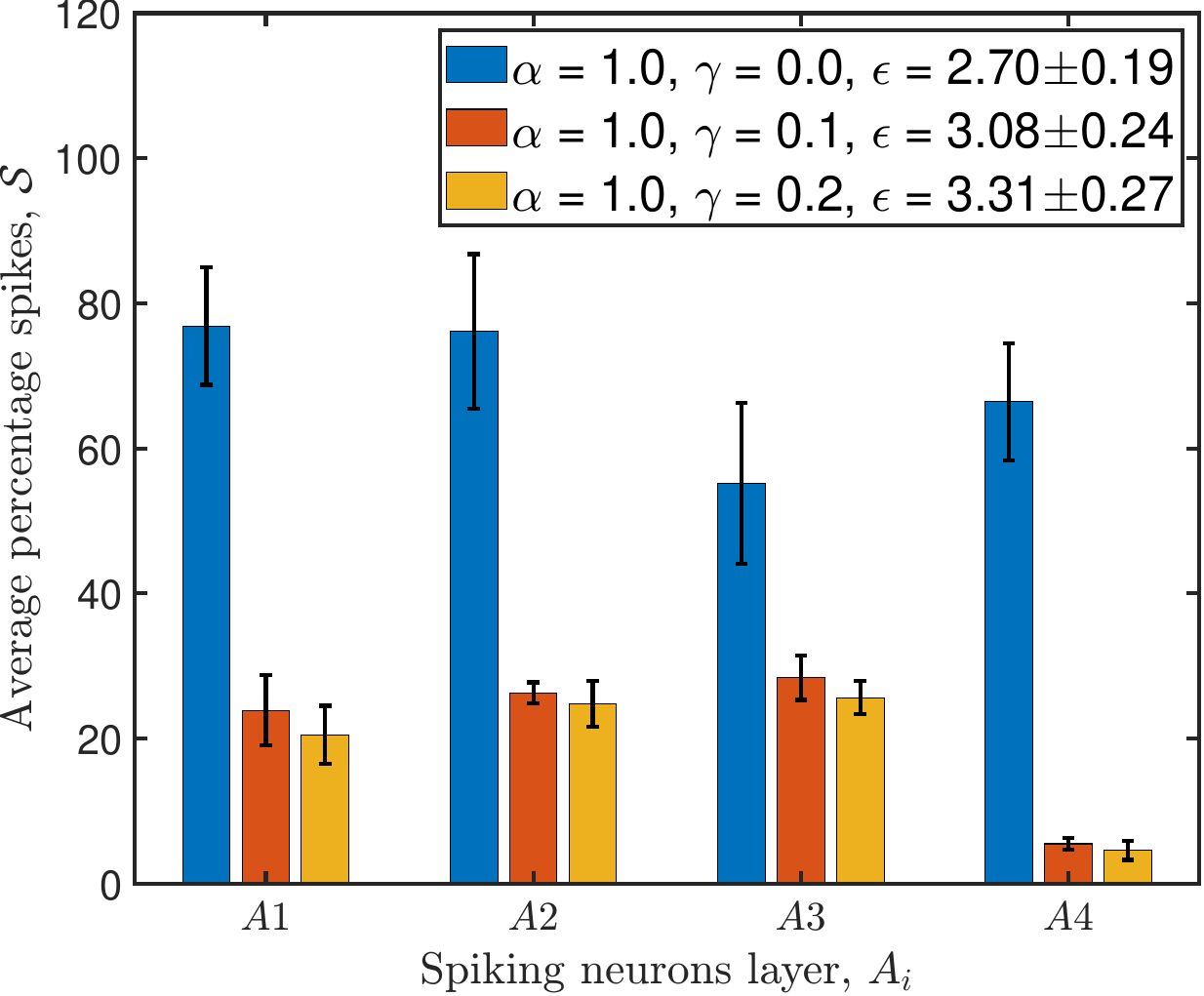}
    \caption{VS-WNO with GeLU activation}
\end{subfigure}    
\caption{Average percentage spikes shown for VS-WNO networks, trained for Burgers' example. The main plotted values are the mean of those observed in five runs of networks, and the standard deviation is shown as error bars. The percentage errors $\epsilon$ observed in different networks are given in the legend. The network performance reported here is for the test dataset.}
\label{fig: burgers spiking activity}
\end{figure}

Fig. \ref{fig: Burgers' VS-WNO OL} further solidifies VS-WNO's (both with GeLU and linear activation) position as the predicted realizations closely follow the ground truth in the test dataset.
Fig. \ref{fig: burgers spiking activity} shows the spiking activity observed in VS-WNO networks. It also shows a reduction in spiking activity when the proposed spiking loss function $L_s$ is used. Note that the parameters set of $\alpha = 1$ and $\gamma = 0$ corresponds to vanilla loss function $L_b$. It can be observed that the increase in percentage error $\epsilon$ from $2.87\pm0.10$ to $3.76\pm0.18$ for VS-WNO-L networks and from $2.70\pm0.19$ to $3.31\pm0.27$ for VS-WNO- GeLU networks is marginal, whereas the spikes produced reduce significantly. The spiking activity saw a reduction as high as $\sim 70\%$ (refer to first spiking layer activity) for VS-WNO-L networks and as high as $\sim 93\%$ (refer to first spiking layer activity) for VS-WNO-GeLU networks.  

\subsection{Example 2: Allen Cahn equation}
The second example deals with a two-dimensional Allen Cahn equation with periodic boundary conditions, defined as,
\begin{equation}
\begin{gathered}
    \partial_t u(x,y,t) = \epsilon\Delta u(x,y,t)+u(x,y,t)-u(x,y,t)^3,\\
    x,y\in(0,3), t\in[0,20],
\end{gathered}
\end{equation}
where $\epsilon = 10^{-3}$ is a real positive constant. The initial condition $u(x,y,0)$ is generated using a Gaussian random field with kernel $\kappa(x,y) = (\pi^2(x^2+y^2)+225)^{1/2}$. The dataset for this example is taken from \cite{tripura2023wavelet}, and the same is discretized on a $x\times y$ grid of $43\times43$. The mapping is carried out between $u(x,y,0)$ and $u(x,y,20)$. Fourteen hundred samples are used to train the various networks. The network details are similar to as discussed before, except the learning rate for LIF-WNO-RE-10 is taken as $10^{-4}$.

\begin{figure}[ht!]
    \centering
    \includegraphics[width = 0.80\textwidth]{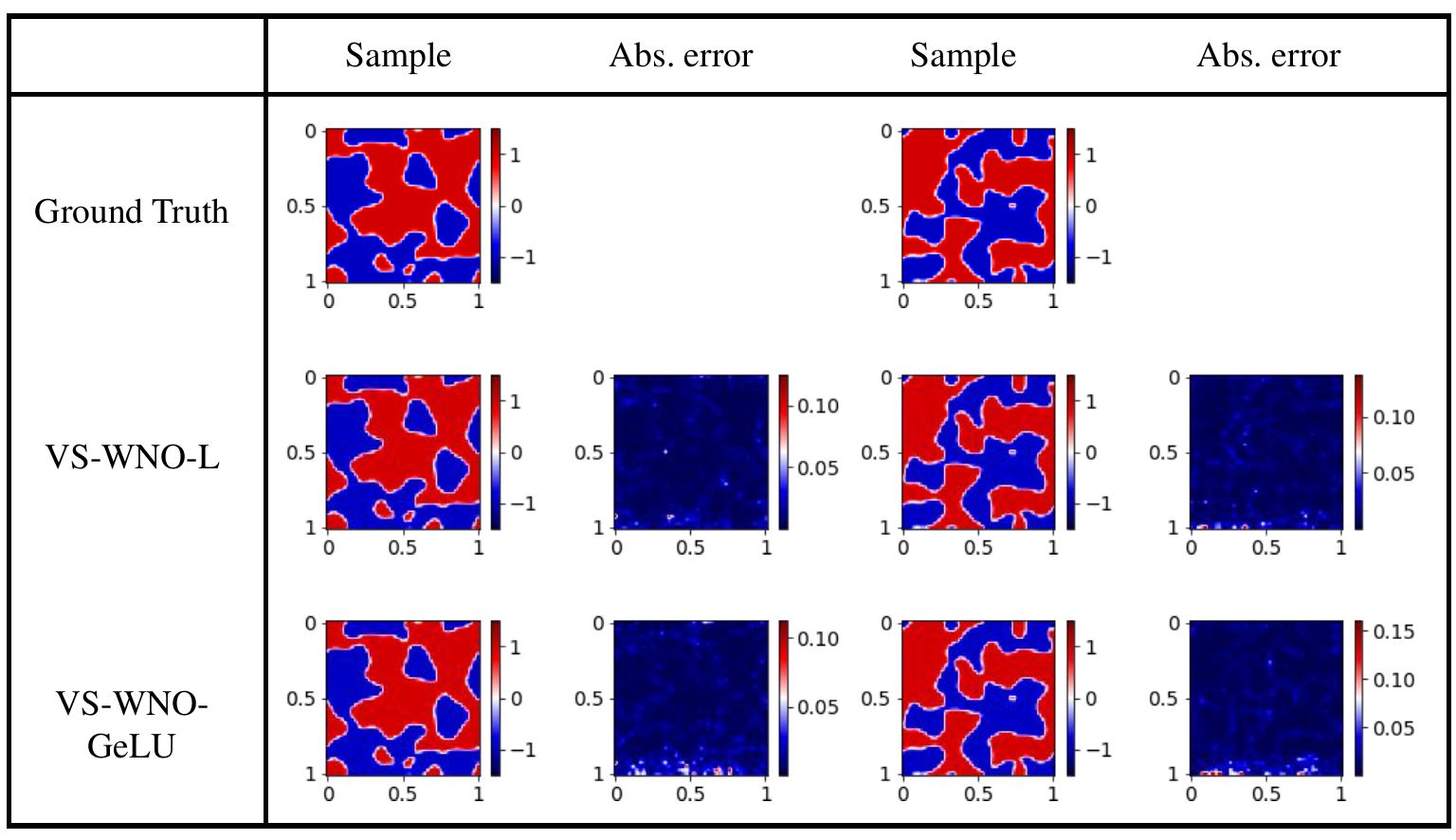}
    \caption{Predictions (two samples) using VS-WNO-L and VS-WNO-GeLU networks, compared against the ground truth, for the Allen Cahn example.}
    \label{fig: Allen Cahn OL}
\end{figure}

\begin{figure}[ht!]
\centering
\begin{subfigure}{0.475\textwidth}
    \centering
    \includegraphics[width = \textwidth]{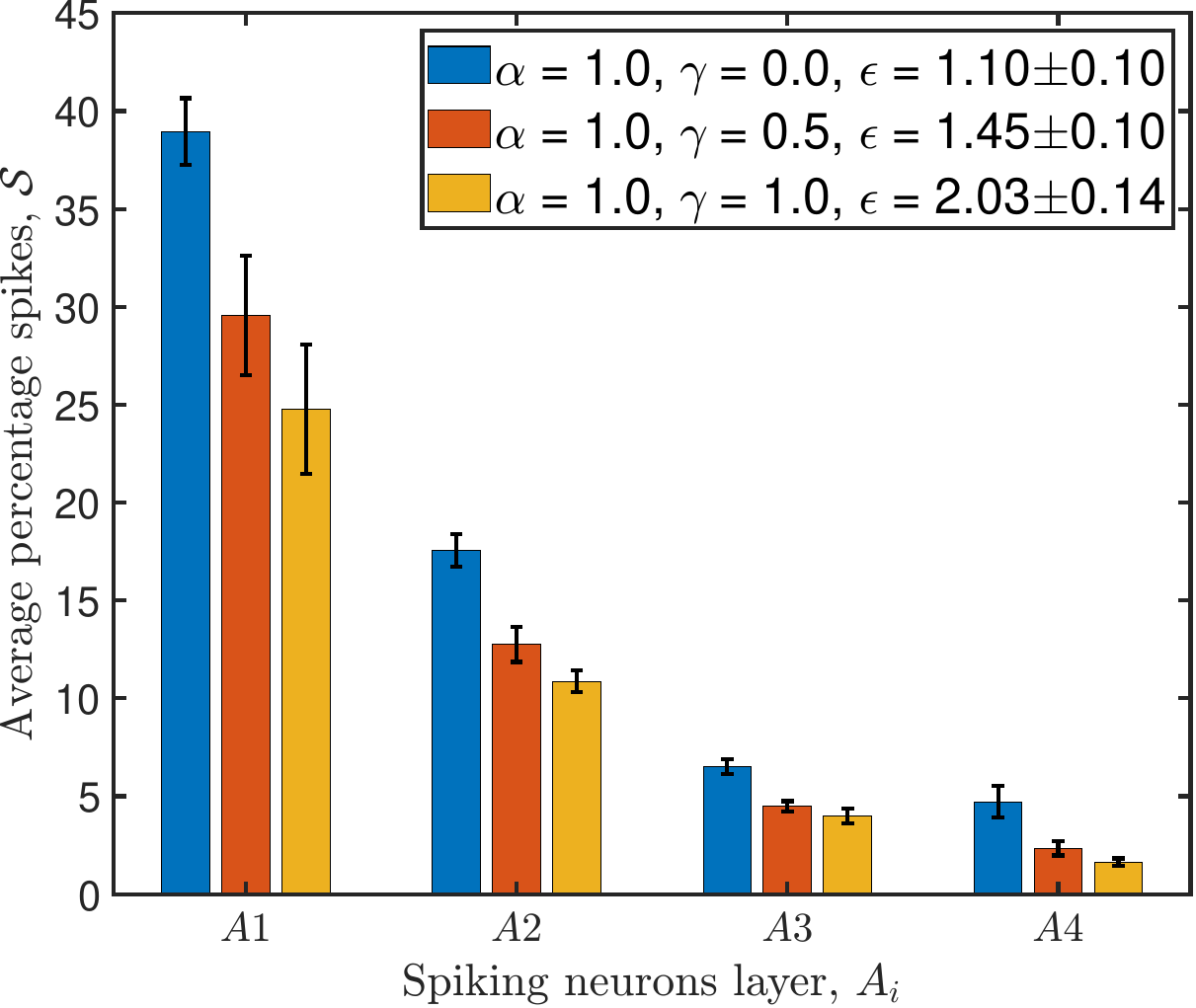}
    \caption{VS-WNO with Linear activation}
\end{subfigure}
\hspace{1em}
\begin{subfigure}{0.475\textwidth}
    \centering
    \includegraphics[width = \textwidth]{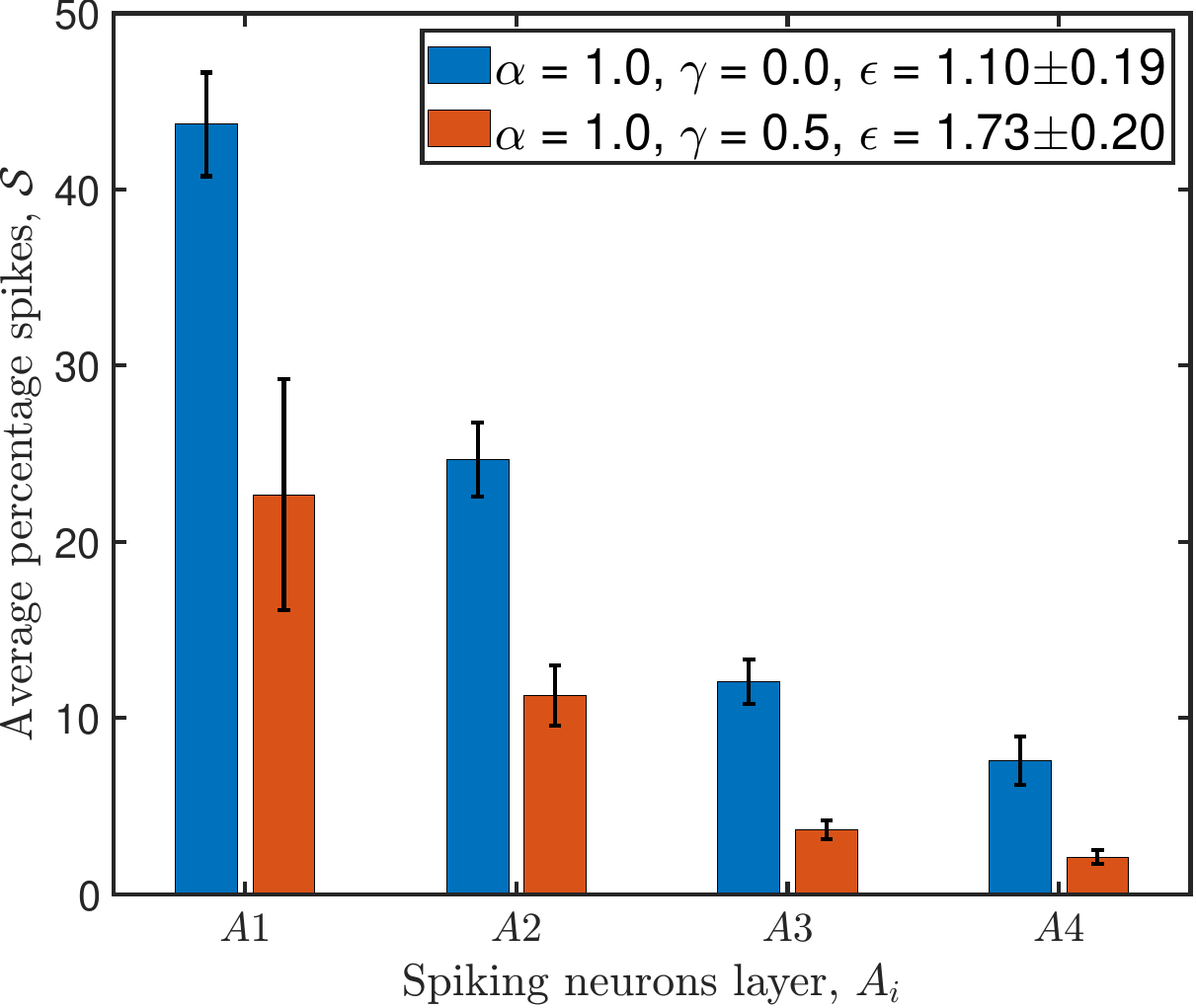}
    \caption{VS-WNO with GeLU activation}
\end{subfigure}    
\caption{Average percentage spikes shown for VS-WNO networks, trained for Allen Cahn example. The main plotted values are the mean of those observed in five runs of networks, and the standard deviation is shown as error bars. The percentage errors $\epsilon$ observed in different networks are given in the legend. The network performance reported here is for the test dataset.}
\label{fig: AC spiking activity}
\end{figure}

The percentage error $\epsilon$ reported in Table \ref{tab: initial results all examples} shows a similar trend, as observed in the previous example. The VS-WNO-L and VS-WNO-GeLU networks converge to ground truth within a single STS. Fig. \ref{fig: Allen Cahn OL} shows predictions for two test samples, and as can be observed, the predictions closely follow the ground truth.
Fig. \ref{fig: AC spiking activity} shows the spiking activity observed for VS-WNO networks along with the drop in spiking activity when using the spiking loss function. Similar to the previous example, a significant drop is observed in the spiking activity of all spiking neurons. For VS-WNO-L networks, the percentage error $\epsilon$ increased from 1.10$\pm$0.10 to $2.03\pm0.14$, while the reduction in spiking activity went as high as $\sim 65\%$ (refer to the last spiking layer). Similarly, for the VS-WNO-GeLU networks, the percentage error increased from 2.70$\pm$0
.19 to 3.31$\pm$0.27, whereas the spiking activity reduced considerably.
\subsection{Example 3: Darcy equation on rectangular domain}
This example discusses a time-independent two-dimensional Darcy flow equation resolved on a rectangular grid. The equation is defined as
\begin{equation}
\begin{gathered}    
    -\nabla(a(x,y)\nabla u(x,y)) = f(x,y),\\
    x,y\in(0,1),
\end{gathered}
\end{equation}
where $u(x,y)$ is the pressure and $a(x,y)$ is the permeability. The source $f(x,y) = 1$ and zero Dirichlet boundary conditions are considered. The permeability $a(x,y)$ is sampled from a Gaussian random field $\psi\mathcal N(0,(-\Delta + 9\mathbb I)^{-2})$, where $\psi$ is mapping which takes a value of 12 on the positive part of real line and takes a value of 3 on the negative side of real line. The dataset is taken from \cite{lu2022comprehensive}, and the mapping is carried out between the permeability $a(x,y)$ and $u(x,y)$. The rectangular grid is discretized on a $85\times85$ grid. A thousand samples are used to train various networks.

\begin{figure}[ht!]
    \centering
    \includegraphics[width = 0.8\textwidth]{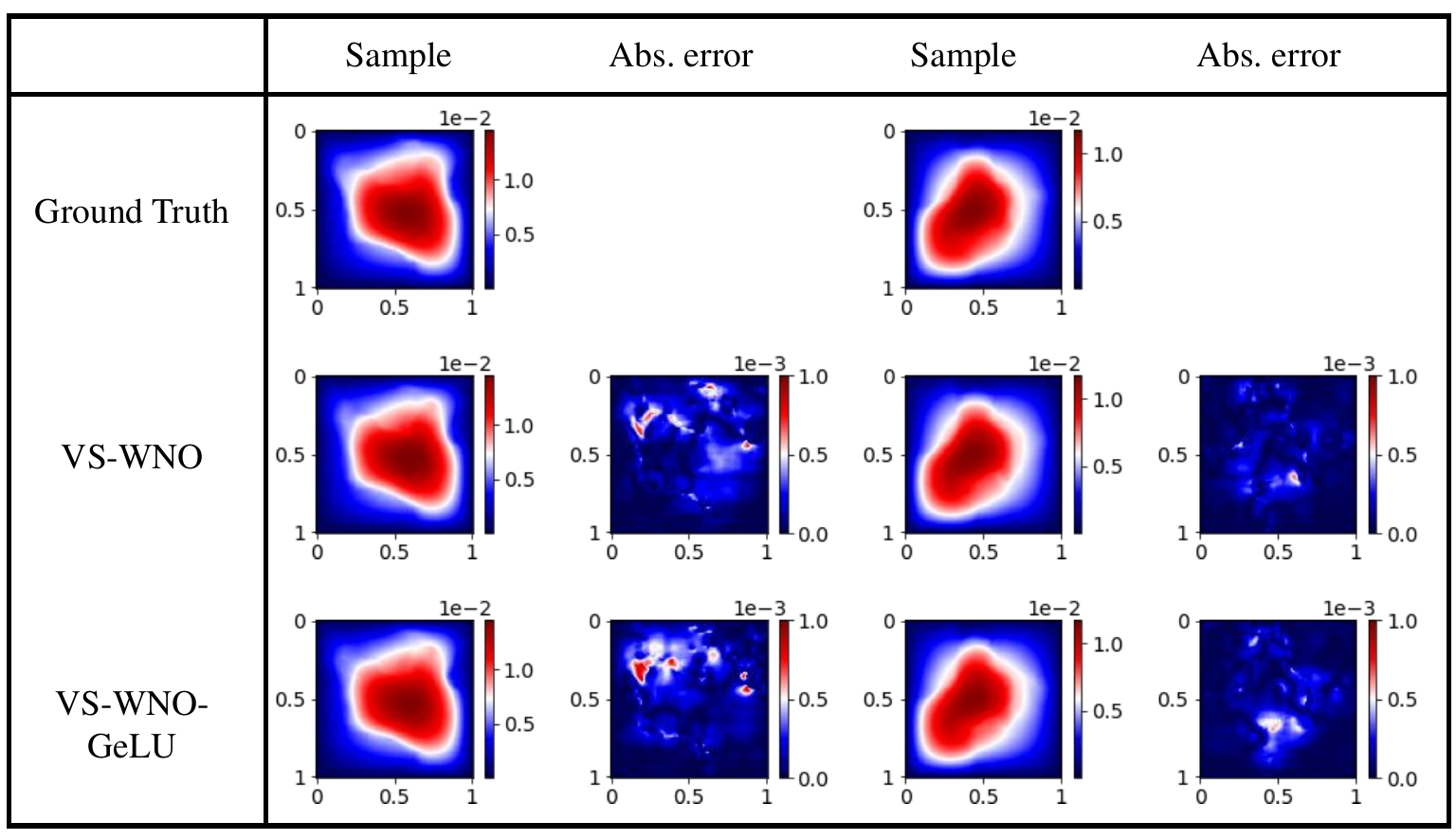}
    \caption{Predictions (two samples) using VS-WNO-L and VS-WNO-GeLU networks, compared against the ground truth, for the Darcy equation on rectangular domain example.}
    \label{fig: Darcy rectangular}
\end{figure}

\begin{figure}[ht!]
\centering
\begin{subfigure}{0.475\textwidth}
    \centering
    \includegraphics[width = \textwidth]{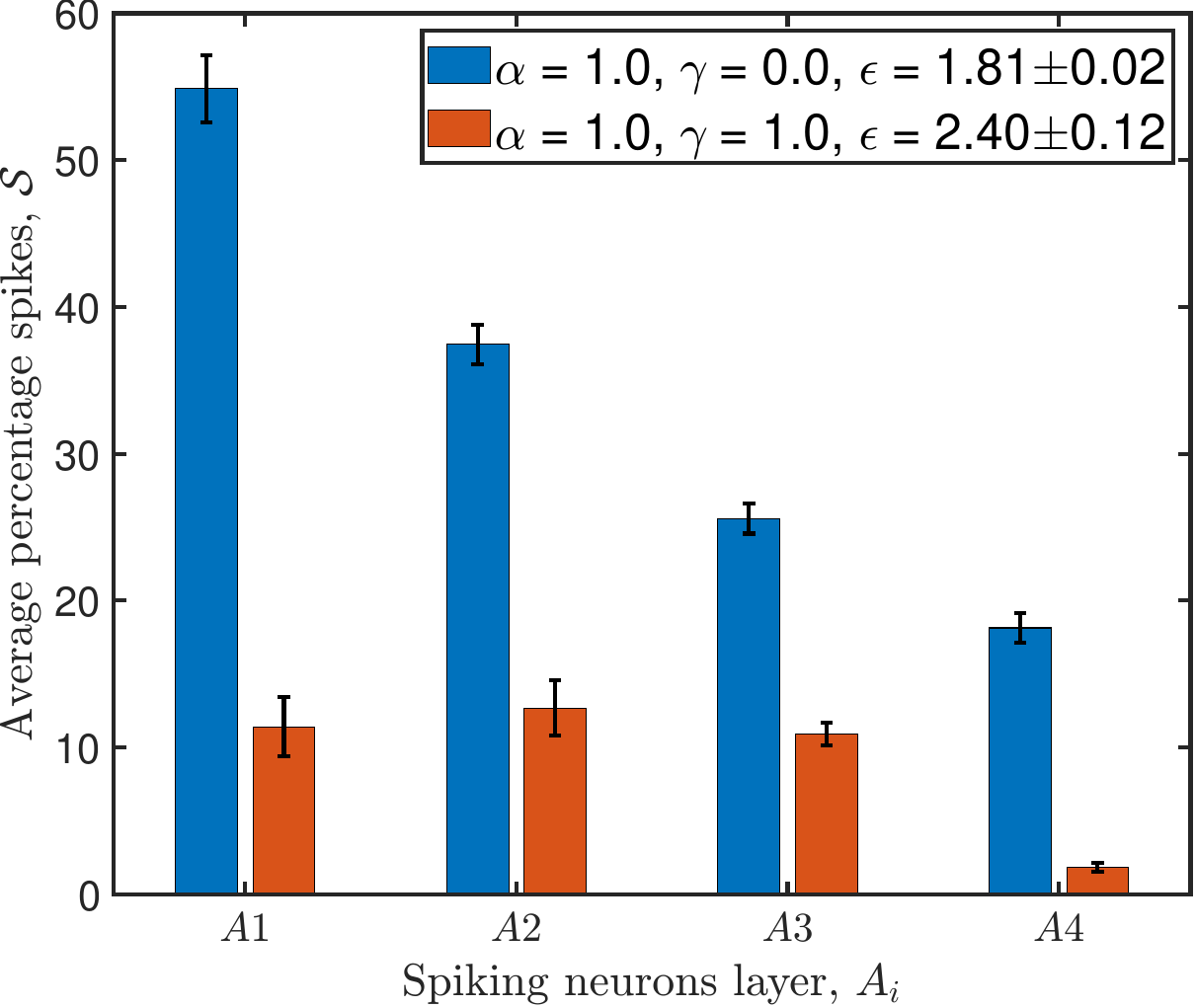}
    \caption{VS-WNO with Linear activation}
\end{subfigure}
\hspace{1em}
\begin{subfigure}{0.475\textwidth}
    \centering
    \includegraphics[width = \textwidth]{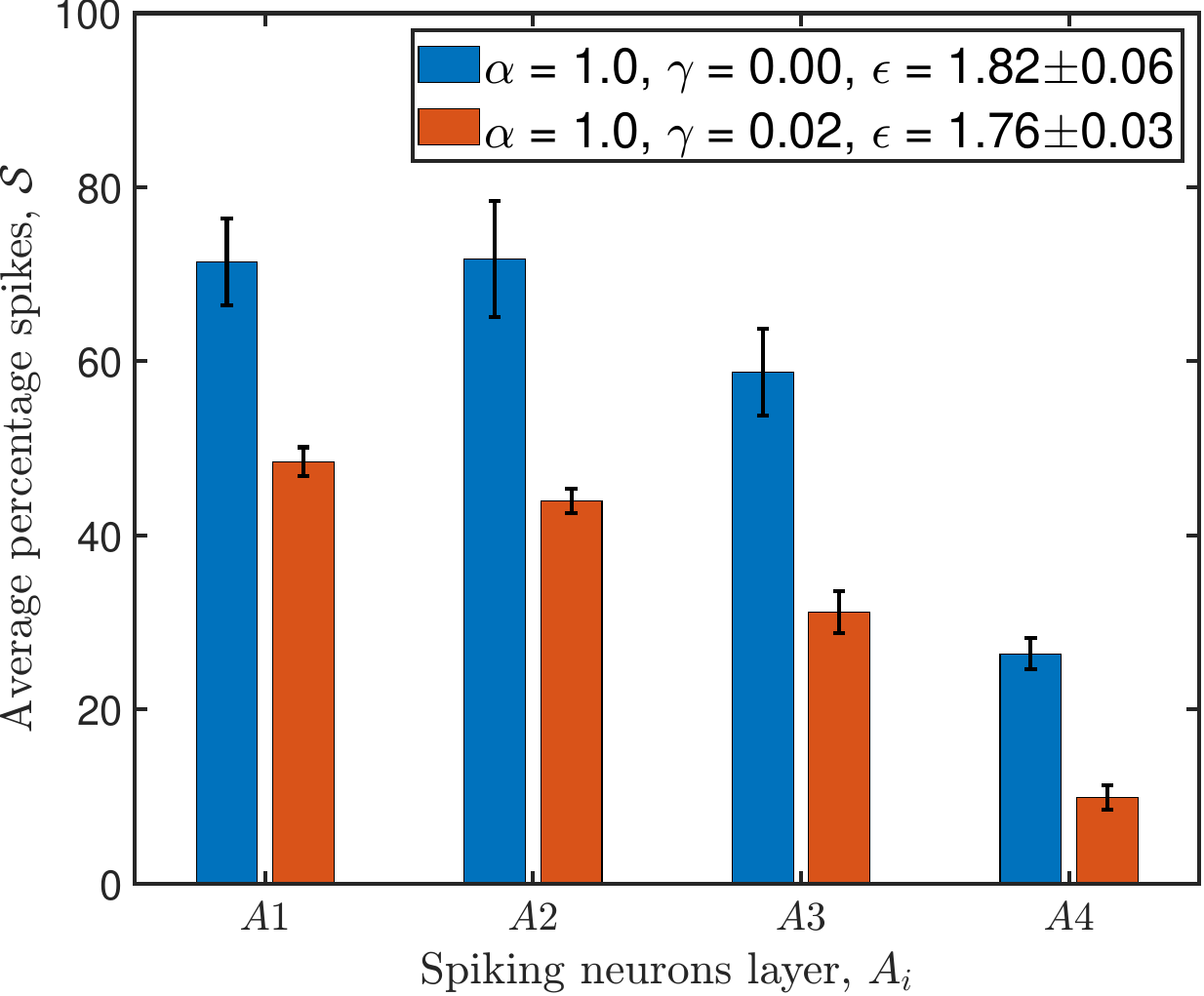}
    \caption{VS-WNO with GeLU activation}
\end{subfigure}    
\caption{Average percentage spikes shown for VS-WNO networks, trained for Darcy equation on rectangular grid example. The main plotted values are the mean of those observed in five runs of networks, and the standard deviation is shown as error bars. The percentage errors $\epsilon$ observed in different networks are given in the legend. The network performance reported here is for the test dataset.}
\label{fig: DRect spiking activity}
\end{figure}

Fig. \ref{fig: Darcy rectangular} shows predictions for two test samples carried out using VS-WNO-L and VS-WNO-GeLU networks. As can be seen, the predicted solutions closely follow the ground truth and the same is reinforced by the observed error values for the test dataset, given in Table \ref{tab: initial results all examples}. 
Fig. \ref{fig: DRect spiking activity} shows the spiking activity observed for VS-WNO networks. The results produced for the VS-WNO-L network show a similar trend as previous examples, wherein a significant drop in spiking activity is observed while the percentage error increased only to $2.40\pm0.12$ for VS-WNO-L networks. For the VS-WNO-GeLU network, while the spiking activity reduced after using the proposed spiking loss function, the percentage error also reduced slightly from 1.82$\pm$0.06 to 1.76$\pm$0.03. The marginal decrease in error can be attributed to better optimization of trainable parameters. 
\subsection{Example 4: Darcy equation on triangular domain}
This example tackles the two-dimensional Darcy problem from the previous example, resolved on a triangular domain $D$, with boundary $\partial D$. The permeability $a(x,y) = 0.1$ and the source $f(x,y) = -1$. The boundary conditions are drawn randomly from a Gaussian random field, and the mapping is carried out between the boundary condition $u(x,y)|_{\partial D}$ and the pressure field $u(x,y)$. The dataset for this case study is taken from \cite{lu2022comprehensive}, and for more details on the data generation, the readers are advised to follow the same. Nineteen hundred samples were used to train various networks.

\begin{figure}[ht!]
    \centering
    \includegraphics[width = 0.8\textwidth]{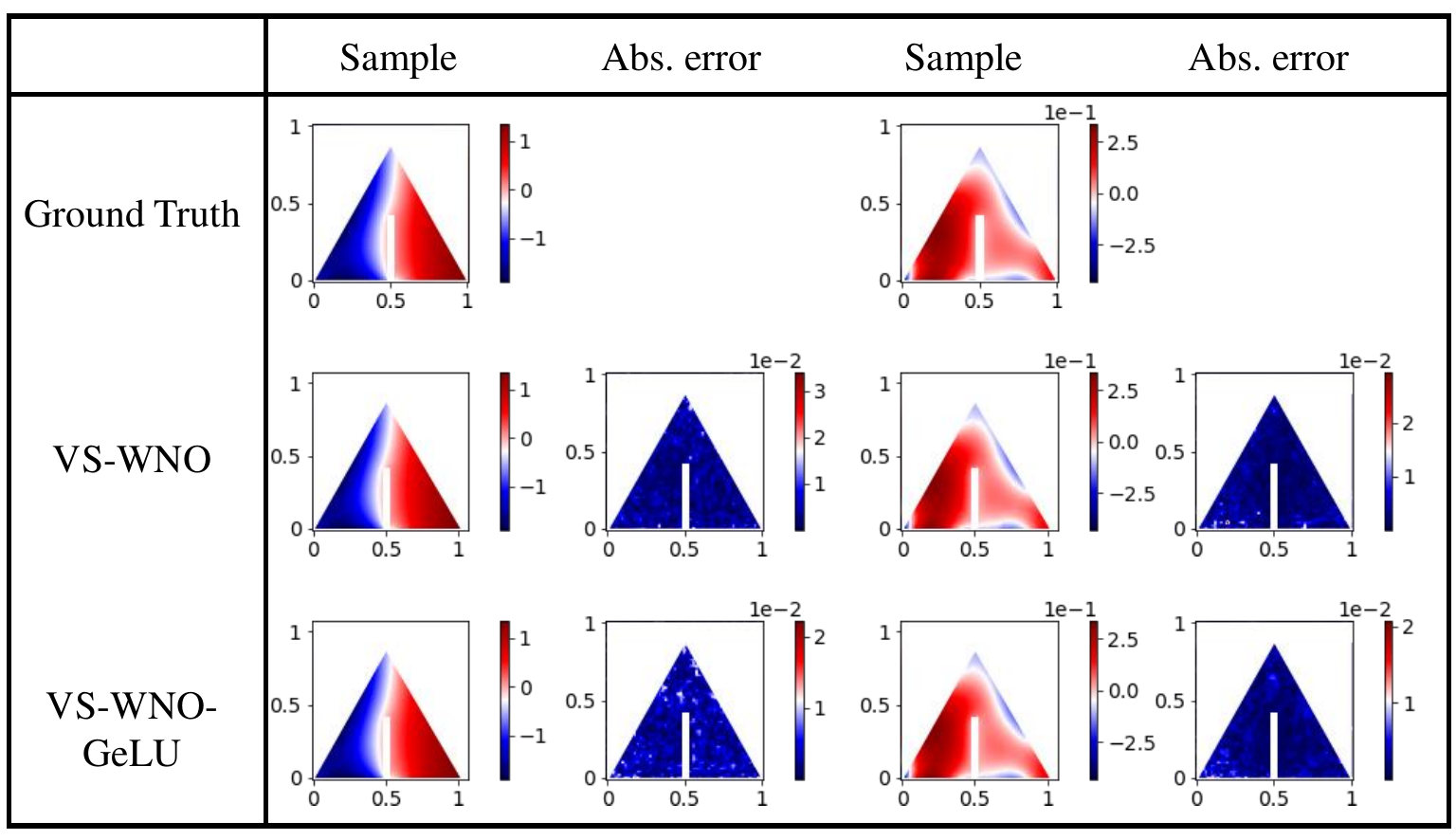}
    \caption{Predictions (two samples) using VS-WNO-L and VS-WNO-GeLU networks, compared against the ground truth, for the Darcy equation on triangular domain example.}
    \label{fig: Darcy triangular}
\end{figure}

\begin{figure}[ht!]
\centering
\begin{subfigure}{0.475\textwidth}
    \centering
    \includegraphics[width = \textwidth]{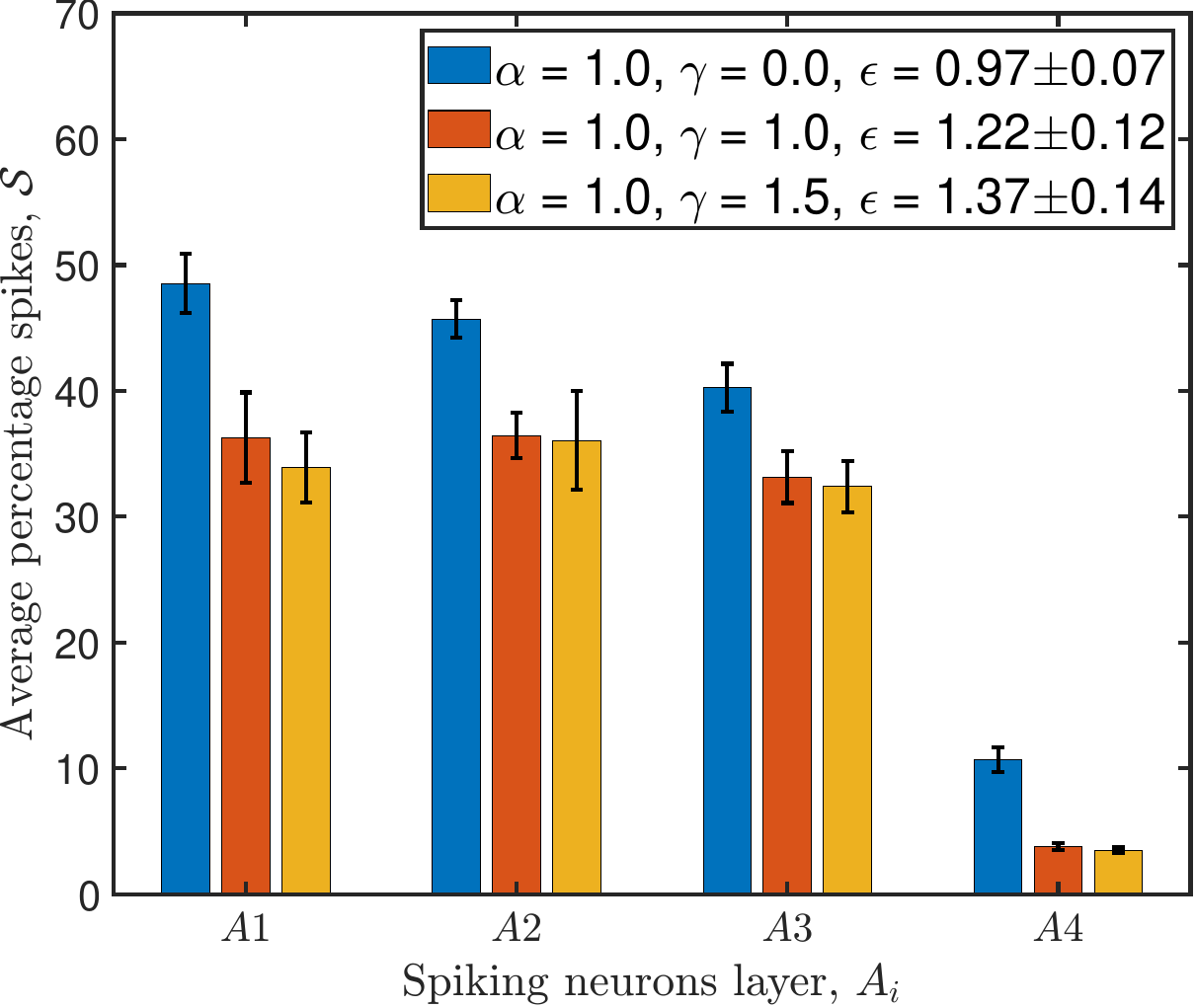}
    \caption{VS-WNO with Linear activation}
\end{subfigure}
\hspace{1em}
\begin{subfigure}{0.475\textwidth}
    \centering
    \includegraphics[width = \textwidth]{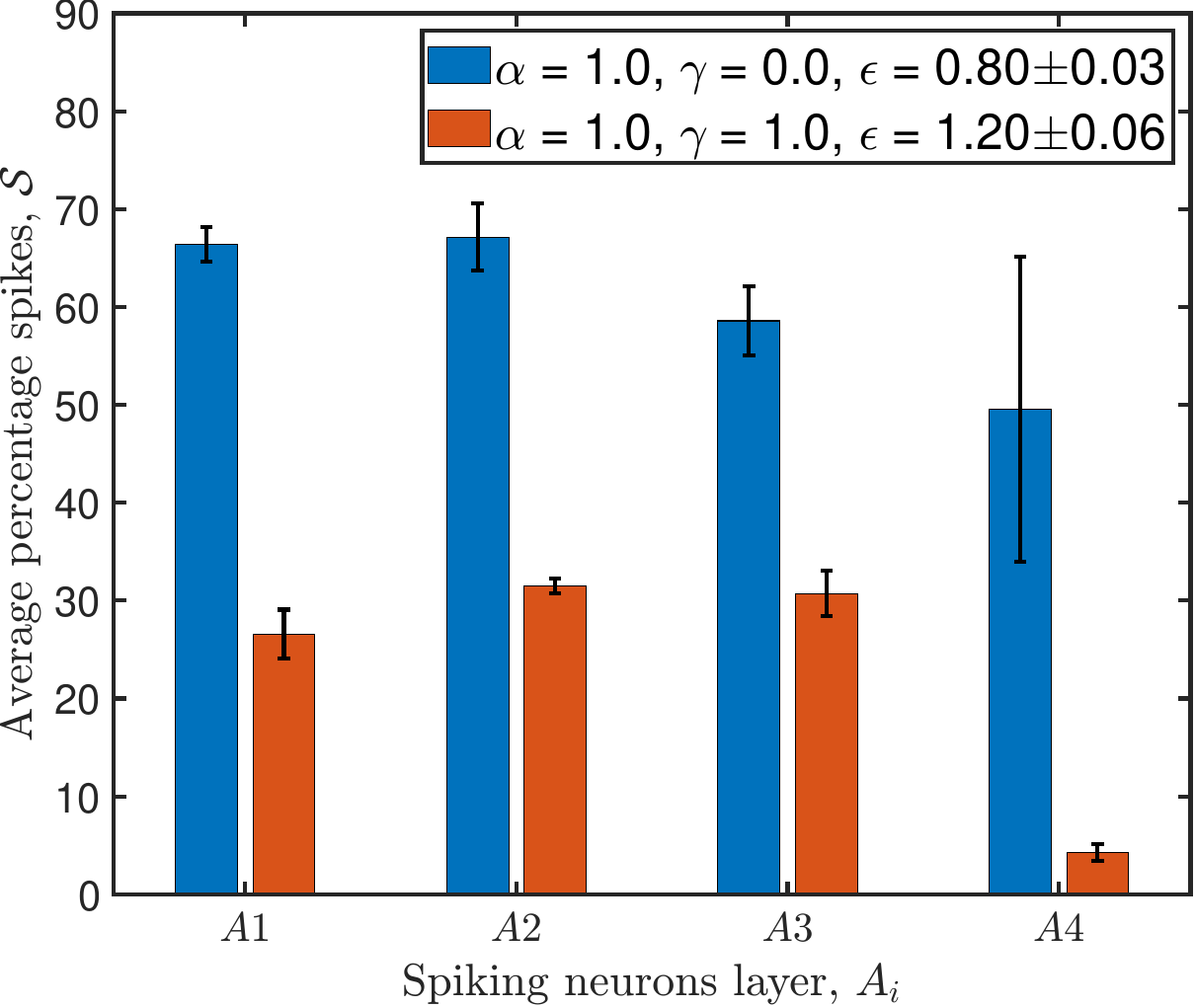}
    \caption{VS-WNO with GeLU activation}
\end{subfigure}    
\caption{Average percentage spikes shown for VS-WNO networks, trained for Darcy equation on triangular grid example. The main plotted values are the mean of those observed in five runs of networks, and the standard deviation is shown as error bars. The percentage errors $\epsilon$ observed in different networks are given in the legend. The network performance reported here is for the test dataset.}
\label{fig: DTri spiking activity}
\end{figure}

\begin{figure}[ht!]
    \centering
    \includegraphics[width = 0.80\textwidth]{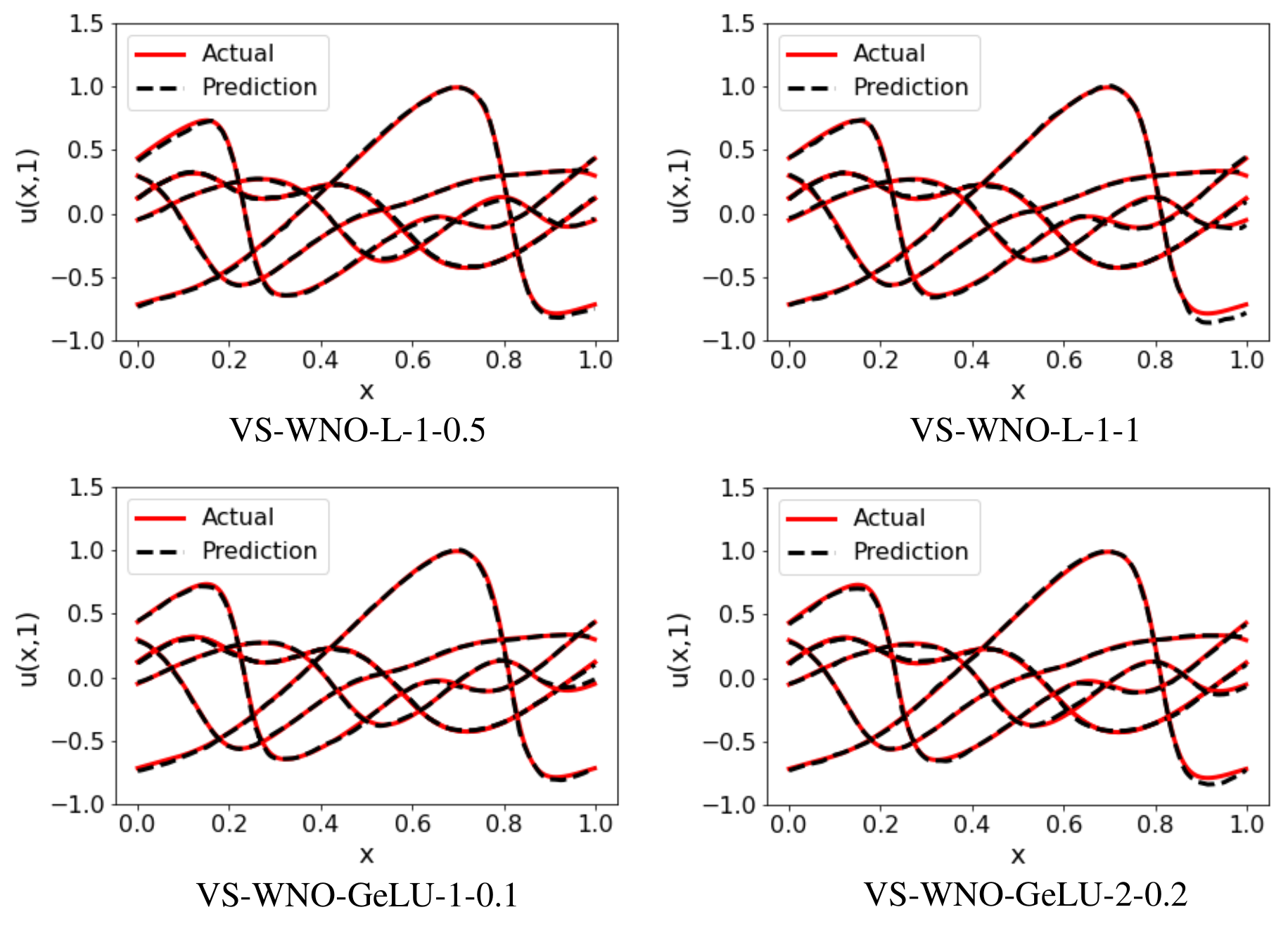}
    \caption{Predictions (five samples) using VS-WNO networks, compared against the ground truth, for the Burgers' example. The networks are trained using the spiking loss function $L_s$. The two numbers at the end of network names, $\#1$ and $\#2$ represent the values for parameters $\alpha$ and $\gamma$, respectively.}
    \label{fig: burger nl}
\end{figure}

\begin{figure}[ht!]
    \centering
    \includegraphics[width = 0.8\textwidth]{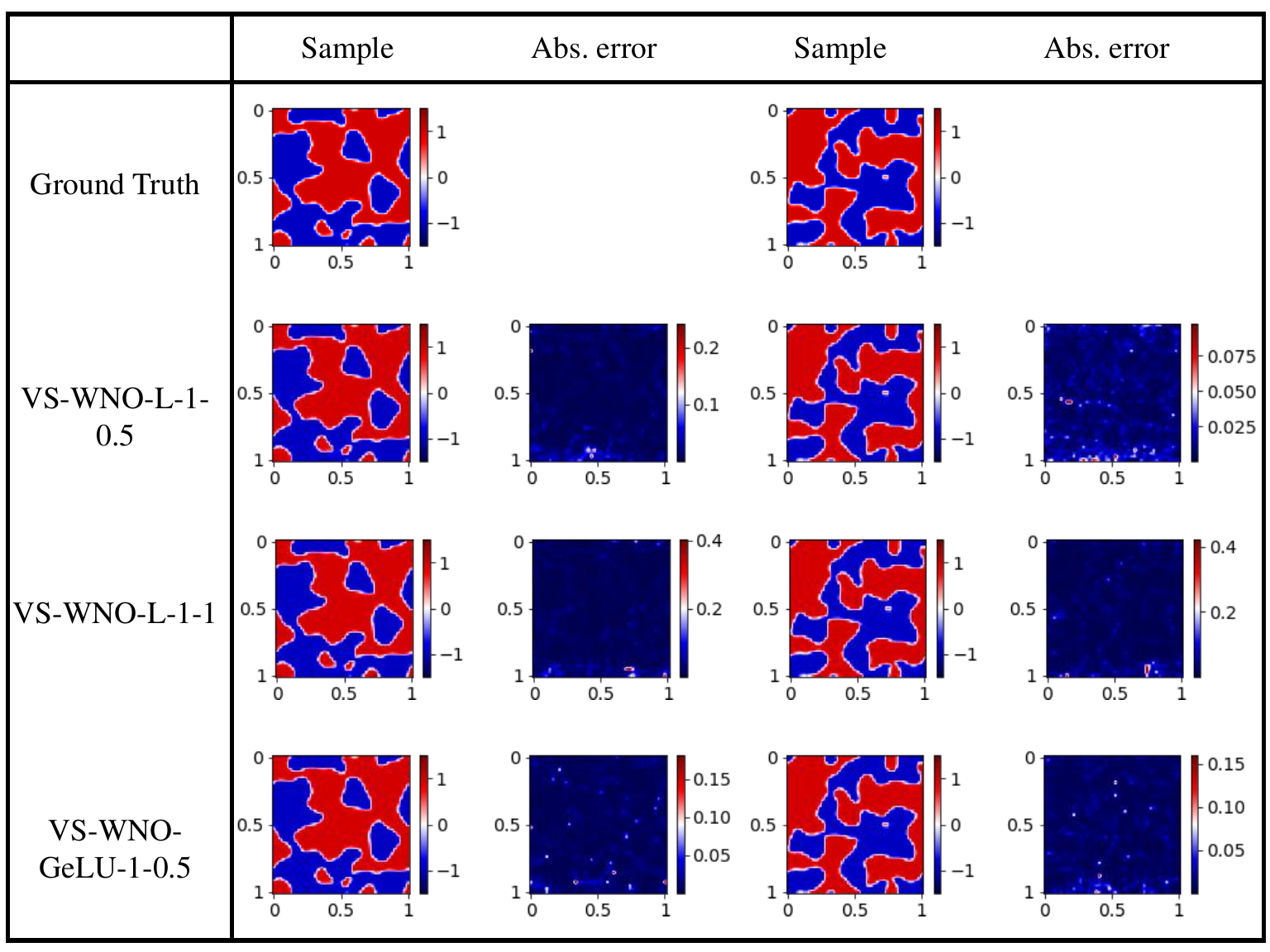}
    \caption{Predictions (two samples) using VS-WNO networks, compared against the ground truth, for the Allen Cahn example. The networks are trained using the spiking loss function $L_s$. The two numbers at the end of network names, $\#1$ and $\#2$ represent the values for parameters $\alpha$ and $\gamma$, respectively.}
    \label{fig: ac nl}
\end{figure}

\begin{figure}[ht!]
    \centering
    \includegraphics[width = 0.8\textwidth]{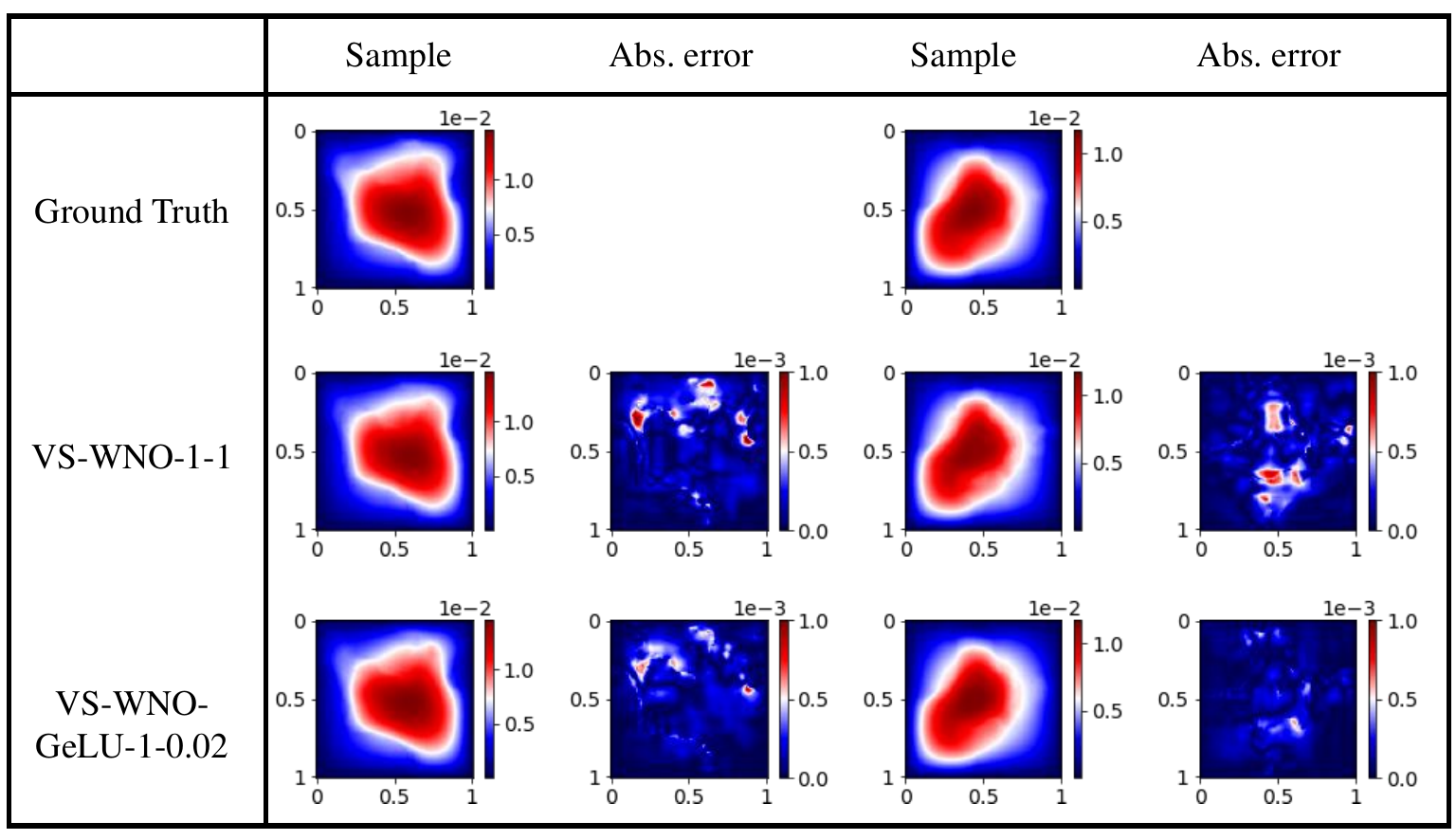}
    \caption{Predictions (two samples) using VS-WNO networks, compared against the ground truth, for the Darcy equation on rectangular domain example. The networks are trained using the spiking loss function $L_s$. The two numbers at the end of network names, $\#1$ and $\#2$ represent the values for parameters $\alpha$ and $\gamma$, respectively.}
    \label{fig: dr nl}
\end{figure}

\begin{figure}[ht!]
    \centering
    \includegraphics[width = 0.8\textwidth]{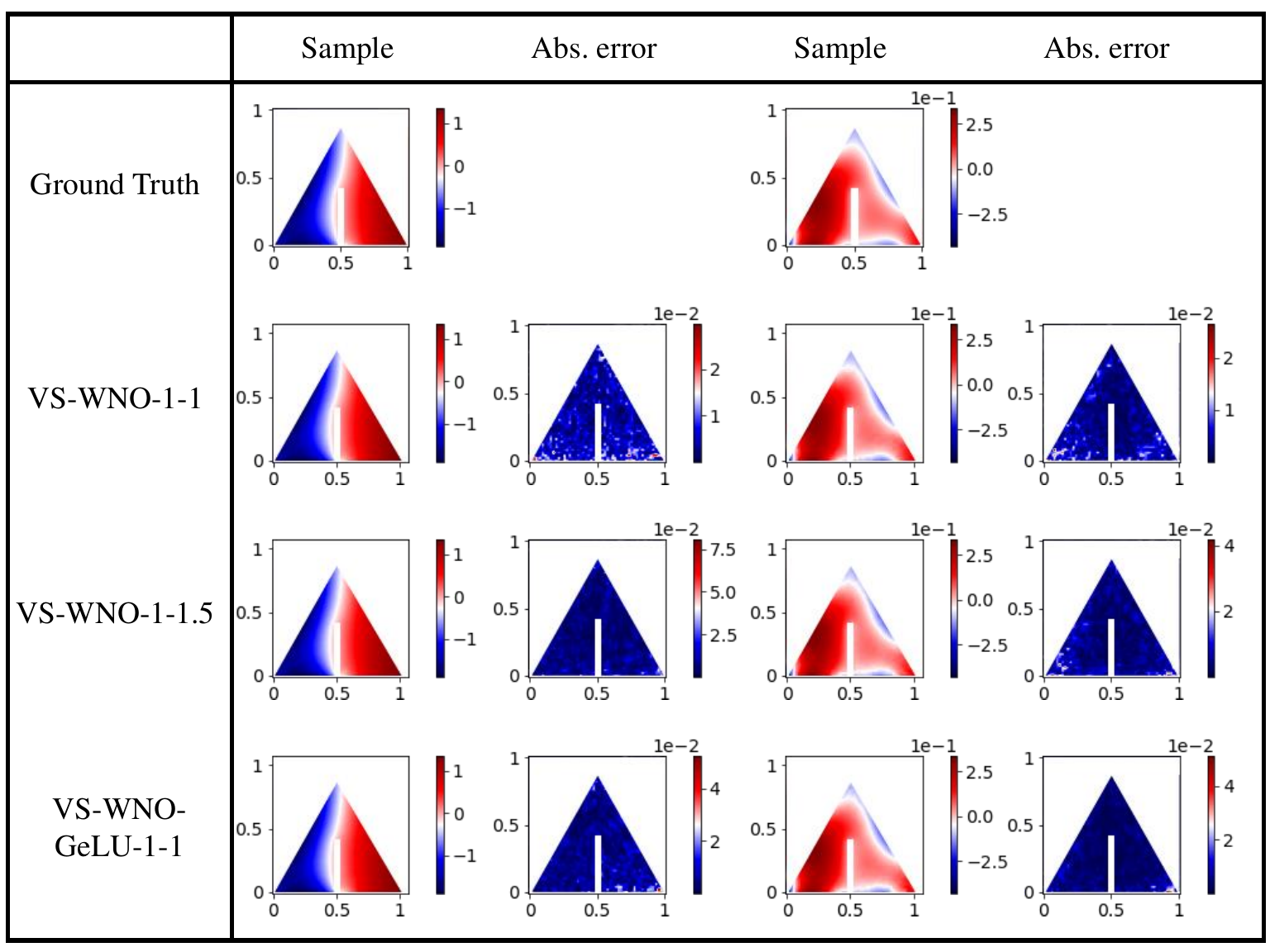}
    \caption{Predictions (two samples) using VS-WNO networks, compared against the ground truth, for the Darcy equation on triangular domain example. The networks are trained using the spiking loss function $L_s$. The two numbers at the end of network names, $\#1$ and $\#2$ represent the values for parameters $\alpha$ and $\gamma$, respectively.}
    \label{fig: dt nl}
\end{figure}

The results for the current example follow the same trend as previous examples, as confirmed by the error values given in Table \ref{tab: initial results all examples} and the predictions shown in Fig. \ref{fig: Darcy triangular}.
Fig. \ref{fig: DTri spiking activity} shows the spiking activity observed in VS-WNO networks for the current example. It can be observed that for the VS-WNO-GeLU network, a significant drop in spiking activity is observed while the increase in percentage error is marginal, from 0.80$\pm$0.03 to 1.20$\pm$0.06. For the VS-WNO-L networks, although not as huge, a drop in spiking activity is observed, and the maximum values observed are well below 50\%. The increase in error is also minuscule, with the increased percentage error being 1.37$\pm$0.14.
Figs. \ref{fig: burger nl}-\ref{fig: dt nl} show predictions for Burgers', Allen Cahn, Darcy's equation on a rectangular grid, and Darcy's equation on a triangular grid, respectively. The training here is carried out using the proposed spiking loss function, and comparisons have been drawn against the ground truth.  As can be seen, despite reduced spiking activity (refer Figs. \ref{fig: burgers spiking activity}, \ref{fig: AC spiking activity}, \ref{fig: DRect spiking activity} and \ref{fig: DTri spiking activity}), the predictions, still follow the ground truth closely and give a good approximation for the same. 
\subsection{Fixed VSN parameters}
In this section, we explore the performance of VS-WNO and the spiking loss function for cases where the leakage and threshold parameters of VSN in VS-WNO are not trained but rather are considered hyperparameters. Now, to initialize VSN neurons in a layer, each VSN is given a leakage and threshold, sampled randomly from a uniform distribution $\mathcal U[0,1]$. Tables \ref{tab: VS-WNO spiking activity no train} and \ref{tab: VS-WNO-GeLU spiking activity no train} show the percentage error observed in Burgers' example and Allen Cahn's example, for the case when the VSN parameters are considered fixed. As can be seen, the VS-WNO networks still converge to ground truth quite well. Also, when the spiking loss function $L_s$ is used while training the network, a reduction in spiking activity is observed, with minimal effect on observed percentage error. As discussed earlier, this may be because the other trainable parameters of the network are being tuned such that the information flowing through the network crosses the threshold of any particular VSN only sparingly.   
\begin{table}[ht!]
    \caption{Spiking activity $\mathcal S$ and percentage normalized $L^2$ values observed in VS-WNO networks for various examples. The VSN parameters are treated as fixed and are initialized randomly.}
    \begin{subtable}[ht!]{\textwidth}
    \caption{Spiking activity $\mathcal S$ and percentage normalized $L^2$ values observed in VS-WNO-L network for various examples.}    \centering
    \begin{tabular}{ccccccccccccccc}
         \toprule
          \multicolumn{7}{c}{Burgers'} && \multicolumn{7}{c}{Allen Cahn} \\\cmidrule{1-7}\cmidrule{9-15}
          
          \multirow{2}{*}{$\alpha$} & \multirow{2}{*}{$\gamma$} & \multirow{2}{*}{$\epsilon$ (\%)} & \multicolumn{4}{c}{percentage spikes, $\mathcal S$} &&
          
          \multirow{2}{*}{$\alpha$} & \multirow{2}{*}{$\gamma$} & \multirow{2}{*}{$\epsilon$ (\%)} & \multicolumn{4}{c}{percentage spikes, $\mathcal S$}\\\cmidrule{4-7}\cmidrule{12-15}
         
         &&& $A1$ & $A2$ & $A3$ & $A4$&&&&& $A1$ & $A2$ & $A3$ & $A4$\\\cmidrule{4-7}\cmidrule{12-15}
         
          1 & 0 & $\underset{\pm 0.15}{3.28}$ & $\underset{\pm 2.53}{42.18}$ & $\underset{\pm 2.62}{32.35}$ & $\underset{\pm 1.91}{19.75}$ & $\underset{\pm 0.45}{2.46}$ &&
          
          1 & 0 & $\underset{\pm 0.24}{1.40}$ & $\underset{\pm 1.85}{44.44}$ & $\underset{\pm 0.68}{21.85}$ & $\underset{\pm 0.66}{7.34}$ & $\underset{\pm 1.01}{5.97}$\\[0.9em]
         
          1 & 0.5 & $\underset{\pm 0.13}{3.66}$ & $\underset{\pm 2.71}{20.73}$ & $\underset{\pm 3.03}{25.09}$ & $\underset{\pm 0.91}{20.34}$ & $\underset{\pm 0.45}{2.77}$ &&
          
          1 & 0.5 & $\underset{\pm 0.14}{1.28}$ & $\underset{\pm 3.19}{32.53}$ & $\underset{\pm 0.56}{16.74}$ & $\underset{\pm 0.11}{6.51}$ & $\underset{\pm 0.79}{5.44}$\\[0.9em]
         
          1 & 1 & $\underset{\pm 0.28}{3.92}$ & $\underset{\pm 4.00}{14.36}$ & $\underset{\pm 2.42}{20.63}$ & $\underset{\pm 1.35}{18.72}$ & $\underset{\pm 0.53}{2.44}$ &&
          
          1 & 1 & $\underset{\pm 0.35}{1.39}$ & $\underset{\pm 4.85}{27.22}$ & $\underset{\pm 1.98}{15.10}$ & $\underset{\pm 0.29}{6.24}$ & $\underset{\pm 0.64}{5.79}$\\
          \bottomrule
         \end{tabular}
    \label{tab: VS-WNO spiking activity no train}
    \end{subtable}
    
    \begin{subtable}[ht!]{\textwidth}
    \vspace{1em}
    \caption{Spiking activity $\mathcal S$ and percentage normalized $L^2$ values observed in VS-WNO-GeLU network for various examples.}    
    \centering
    \begin{tabular}{ccccccccccccccc}
         \toprule
          \multicolumn{7}{c}{Burgers'} && \multicolumn{7}{c}{Allen Cahn} \\\cmidrule{1-7}\cmidrule{9-15}
          
          \multirow{2}{*}{$\alpha$} & \multirow{2}{*}{$\gamma$} & \multirow{2}{*}{$\epsilon$ (\%)} & \multicolumn{4}{c}{percentage spikes, $\mathcal S$} &&
          
          \multirow{2}{*}{$\alpha$} & \multirow{2}{*}{$\gamma$} & \multirow{2}{*}{$\epsilon$ (\%)} & \multicolumn{4}{c}{percentage spikes, $\mathcal S$}\\\cmidrule{4-7}\cmidrule{12-15}
         
         &&& $A1$ & $A2$ & $A3$ & $A4$&&&&& $A1$ & $A2$ & $A3$ & $A4$\\\cmidrule{4-7}\cmidrule{12-15}
         
          1 & 0 & $\underset{\pm 0.19}{3.31}$ & $\underset{\pm 0.91}{39.73}$ & $\underset{\pm 0.95}{29.53}$ & $\underset{\pm 2.26}{17.62}$ & $\underset{\pm 0.49}{3.04}$ &&
          
          1 & 0 & $\underset{\pm 0.10}{0.74}$ & $\underset{\pm 2.49}{44.61}$ & $\underset{\pm 1.78}{28.03}$ & $\underset{\pm 1.91}{11.84}$ & $\underset{\pm 1.09}{9.81}$\\[0.9em]
         
          1 & 0.1 & $\underset{\pm 0.22}{3.96}$ & $\underset{\pm 1.78}{8.39}$ & $\underset{\pm 3.12}{24.10}$ & $\underset{\pm 3.20}{19.82}$ & $\underset{\pm 0.53}{3.14}$ &&
          
          1 & 0.5 & $\underset{\pm 0.17}{0.99}$ & $\underset{\pm 6.12}{26.48}$ & $\underset{\pm 1.85}{15.88}$ & $\underset{\pm 1.13}{7.12}$ & $\underset{\pm 2.23}{8.09}$\\[0.9em]
         
          2 & 0.2 & $\underset{\pm 0.10}{3.99}$ & $\underset{\pm 1.71}{9.21}$ & $\underset{\pm 3.51}{24.16}$ & $\underset{\pm 1.90}{19.78}$ & $\underset{\pm 0.55}{3.21}$ &&
          
          & & & & & & \\
          \bottomrule
          \end{tabular}
    \label{tab: VS-WNO-GeLU spiking activity no train}
    \end{subtable}
\end{table}
\section{Conclusion}\label{section: conclusion}
In this paper, we introduce a variable spiking neural operator (VS-WNO) for solving computational mechanics problems. In scientific computing, we often come across problems involving complex PDEs, and recent strides in AI algorithms, like operator learning algorithms, have shown great promise in tackling such problems. The idea behind introducing a spiking operator learning scheme is to tailor AI algorithms such that they can make the transition from a perspective to a practical solution. 
To implement VS-WNO, we take the help of the Variable Spiking Neuron (VSN), tailored for regression tasks. VSN promotes communication sparsity and excels in handling continuous activations, creating an amalgamation of LIF spiking neurons and artificial neurons. Within the VS-WNO architecture, VSN layers replace conventional activations. Additionally, we propose a spiking loss function to control network spiking activity.

Various examples are considered to test the proposed framework, including both the one-dimensional and two-dimensional PDEs. The key observations from the results produced are as follows, 
\begin{itemize}
    \item The proposed VS-WNO outperforms WNO utilizing LIF neurons (a popular spiking neuron) in all the examples, and its performance is comparable to that of vanilla WNO despite utilizing sparse communication.
    \item The proposed VS-WNO converges to ground truth within a single STS and produces good results with direct inputs.
    \item VS-WNO with linear activation is at par with VS-WNO with GeLU activation. However, VS-WNO with linear activation will be cheaper from a resource consumption point of view since the computations for continuous activations will not be required.
    \item The spiking activity observed in VS-WNO drops significantly with the use of proposed spiking loss $L_s$.
    \item The proposed spiking loss function is able to reduce spiking activity even if the parameters of VSNs in VS-WNO are treated as fixed. 
    \item The increase in error because of increased sparsity, when using the spiking loss function, is marginal, and the results produced still follow the ground truth well.
\end{itemize}
Another benefit of using VS-WNO, observed at the time of compiling results, was the predictions don't require additional smoothing or post-processing (as done in \cite{kahana2022function,zhang2023artificial}) despite using a threshold function within the architecture. The authors would like to clarify here that energy consumed by VS-WNO, utilizing VSN, lies somewhere in between WNO utilizing LIF neurons and WNO utilizing artificial neurons.
This is because in LIF neurons, reduced communication, as well as elimination of multiplication operations, contribute to energy saving, while in VSN, sparse communication is the primary source of energy consumption.

Having said all this, the authors here would like to note that despite VS-WNO's excellent performance in regression tasks, there is scope for further improvement in terms of accuracy. Furthermore, the parameters $\alpha$ and $\gamma$ of the proposed spiking loss function, if not selected judiciously, can cause stability issues. Also, the research in the field of regression tasks using spiking networks is still at its nascent stage, and a whole infrastructure needs to be developed to utilize the full benefits of spiking networks. One of the open problems in this realm is to develop a training algorithm that can cater to spiking neural networks without approximations, as made in surrogate backpropagation. We believe that the proposed VS-WNO is a step in the right direction and warrants appropriate consideration.
\section*{Acknowledgment}
SG acknowledges the financial support received from the Ministry of Education, India, in the form of the Prime Minister's Research Fellows (PMRF) scholarship. SC acknowledges the financial support received from the Science and Engineering Research Board (SERB) via grant no. SRG/2021/000467 and from Ministry of Port and Shipping via letter no. ST-14011/74/MT (356529).
\end{document}